\documentclass[10pt,twocolumn,letterpaper]{article}

\usepackage{iccv}
\usepackage{times}
\usepackage{epsfig}
\usepackage{graphicx}
\usepackage{amsmath}
\usepackage{amssymb}
\usepackage{animate}

\usepackage{booktabs}

\usepackage{algorithm}
\usepackage{algorithmic}
\usepackage{xspace}

\usepackage[accsupp]{axessibility}  

\usepackage{xfrac}

\usepackage[most]{tcolorbox}
\usepackage{float}
\usepackage{tabularx}
\newcolumntype{?}{!{\vrule width 2pt}}

\usepackage[pagebackref=true,breaklinks=true,letterpaper=true,colorlinks,bookmarks=false]{hyperref}

\iccvfinalcopy 

\def\iccvPaperID{7102} 

\newcommand{\methodName}{RIVER\xspace}
\newcommand{\methodNameExtended}{Random frame conditioned flow Integration for VidEo pRediction\xspace}
\def\kitti{0}
\newcommand{\warm}{warm-start sampling\xspace}

\newcommand{\aram}[1]{\textcolor{black}{#1}\xspace}
\newcommand{\sepehr}[1]{\textcolor{black}{#1}\xspace}

\ificcvfinal\fi

\begin{document}

\title{Efficient Video Prediction via Sparsely Conditioned Flow Matching}

\newcommand{\footremember}[2]{%
   \thanks{#2}
    \newcounter{#1}
    \setcounter{#1}{\value{footnote}}%
}
\newcommand{\footrecall}[1]{%
    \footnotemark[\value{#1}]%
} 

\author{Aram Davtyan\footremember{eq}{Equal contribution.}, \quad Sepehr Sameni\footrecall{eq}, \quad Paolo Favaro\\ 
Computer Vision Group, Institute of Computer Science, University of Bern, Switzerland\\
{\tt\small \{aram.davtyan, sepehr.sameni, paolo.favaro\}@unibe.ch}
}

\maketitle
\ificcvfinal\thispagestyle{empty}\fi

\begin{abstract}
We introduce a novel generative model for video prediction based on latent flow matching, an efficient alternative to diffusion-based models. 
In contrast to prior work, we keep the high costs of modeling the past during training and inference at bay by conditioning only on a small random set of past frames at each integration step of the image generation process. 
Moreover, to enable the generation of high-resolution videos and to speed up the training, we work in the latent space of a pretrained VQGAN.
Finally, we propose to approximate the initial condition of the flow ODE with the previous noisy frame. This allows to reduce the number of integration steps and hence, speed up the sampling at inference time.
We call our model \methodNameExtended, or, in short, \methodName. We show that \methodName achieves superior or on par performance compared to prior work on 
common
video prediction benchmarks\aram{, while requiring an order of magnitude fewer computational resources.} Project website: \url{https://araachie.github.io/river}.
\end{abstract}


\begin{figure}
    \centering
    \includegraphics[width=\linewidth]{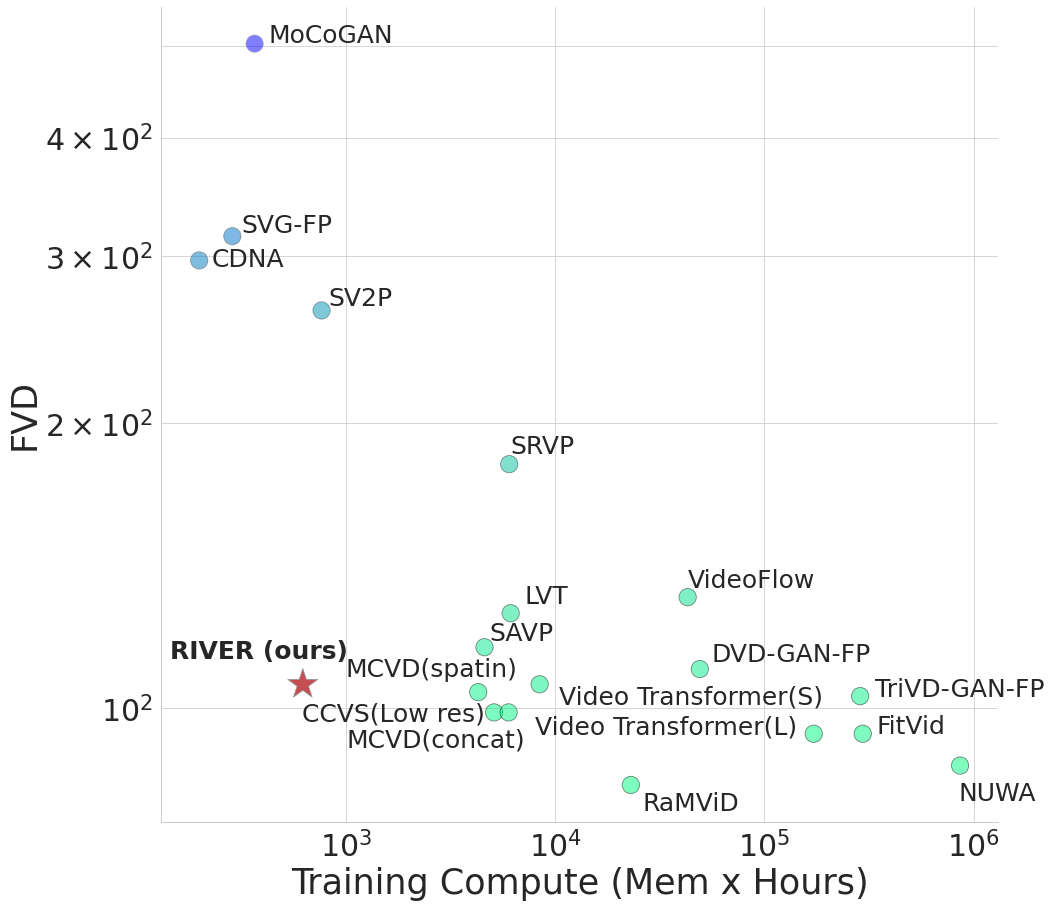}
    \caption{\aram{\emph{\methodName} achieves an ideal trade-off between quality of generated videos and compute needed to train the model. This makes research on video models more easily scalable.}}
    \label{fig:tradeoff}
\end{figure}

\section{Introduction}
\label{sec:intro}

Video prediction, \ie, the task of predicting future frames given past ones, is a fundamental component of an agent that needs to interact with an environment \cite{babaeizadeh2021fitvid}. 
This capability enables planning and advanced reasoning, especially when other agents are in the scene \cite{finn2017deep, finn2016unsupervised, xie2021learning}.
More in general, however, a video prediction model that can generalize to new unseen scenarios needs to implicitly understand the scene, \ie, detect and classify objects, learn how each object moves and interacts, estimate the 3D shape and location of the objects, model the laws of physics of the environment, and so on. In addition to naturally leading to rich and powerful representations of videos, this task does not require any labeling and is thus an excellent candidate for learning from readily available unannotated datasets. 

While the literature in video prediction is by now relatively rich \cite{denton2018stochastic,babaeizadeh2021fitvid, lee2018stochastic}, the quality of the predicted frames has been achieving realistic levels only recently \cite{ voleti2022mcvd,hoppe2022diffusion,harvey2022flexible,yang2022diffusion}. This has been mostly due to the exceptional complexity of this task and the difficulty of training models that can generalize well to unseen (but in-domain) data.

To address these challenges, we propose a novel training procedure for video prediction that is computationally efficient and delivers high quality frame prediction.
One of the key challenges of synthesizing realistic predicted frames is to ensure the temporal consistency of the generated sequence. To this aim, conditioning on as many past frames as possible is a desirable requirement. In fact, with only two past frames it is possible to predict only constant motions at test time, and for general complex motions, such as object interactions (\eg, a ball bouncing off a cube in CLEVRER~\cite{Yi2020CLEVRERCE}), many more frames are needed. However, conditioning on many past frames comes either at the sacrifice of the video quality or at a high computational cost, as shown in Figure~\ref{fig:tradeoff}. In the literature, we see two main approaches to address these issues: 1) models that take a fixed large temporal window of past frames as input and 2) models that compress all the past into a state, such as recurrent neural networks (RNNs) \cite{Babaeizadeh2018StochasticVV,denton2018stochastic,babaeizadeh2021fitvid}.
Fixed window models require considerable memory and computations both during training and at inference time. Although methods such as Flexible Diffusion~\cite{harvey2022flexible} can gain considerable performance by choosing carefully non contiguous past frames, their computational cost still remains demanding.
RNNs also require considerable memory and computations resources at training time, as they always need to feed a sequence from the beginning to learn how to predict the next frame. Moreover, training these models is typically  challenging due to the vanishing gradients.


In the recent growing field of diffusion models for image generation, the 3DiM method \cite{watson2022novel} introduces the idea of a sparse conditioning on randomly chosen scene views during the diffusion process, and showed impressive results in novel view synthesis. In our approach, we adapt this idea to the case of video prediction by also conditioning the generation of the next frame on a randomly chosen sparse set of past frames during the diffusion process. In practice, this is an effective remedy to limit the computational complexity of the model at both training and test time, because the conditioning at each training step is limited to a small set of past frames, but the frame prediction at test time can incorporate an arbitrary number of past frames (also efficiently).
To further speed up the training and generation of videos at test time, we compress videos via VQGAN autoencoding \cite{esser2021taming} and work in the latent space. This design choice has been shown to work well in the case of image generation \cite{rombach2022high} and to enable the efficient generation of images at high resolution.
\sepehr{Unlike other methods that employ a temporal VQGAN~\cite{yan2021videogpt,gupta2022maskvit}, we adopt a per-frame VQGAN approach to minimize the training cost. Additionally, we incorporate a refinement network (more details in section~\ref{sec:vidpred}) to improve the frame quality and to correct any temporal inconsistencies between pairs of frames during post-processing.}
We gain another significant performance boost both in terms of better convergence at training time and in terms of better image quality generation, by adapting Flow Matching \cite{lipman2022flow} to video prediction. The key insight is that it is possible to build an explicit mapping of a noise instance to an image sample, and diffusion models are the result of a specific choice of such mapping, which has been shown experimentally to be sub-optimal \cite{lipman2022flow}.
Finally, to make our method more efficient at inference time, we introduce a \emph{\warm}. In the case of video prediction, the content changes slowly over time. Thus, we expect that a very good guess for the next frame is the current frame itself. Therefore, we propose to speed up the integration of the flow to generate the next frame by starting from a noisy current frame rather than from zero-mean Gaussian noise.
We call our method  \methodNameExtended (\methodName).
We demonstrate \methodName on common video prediction benchmarks and show that it performs on par or better than state of the art methods, while being much more efficient to train. We also show that \methodName can be used for video generation and interpolation and can predict non-trivial long-term object interactions.
In summary, our contributions are the design of a video prediction model that
\begin{enumerate}
    \item Extends flow matching to video prediction;
    \item Is  efficient to train and to use at test time; 
    \item Can be conditioned on arbitrarily many past frames;
    \item Is efficient at test time (via \warm).
\end{enumerate}

\section{Prior work}
\label{sec:prior}

\noindent\textbf{Conventional Methods.} Video prediction models are used to generate realistic future frames of a video sequence based on past frames. Up until recently, most video prediction models relied on Recurrent Neural Networks (RNNs) in the bottleneck of a convolutional autoencoder~\cite{babaeizadeh2021fitvid}. Training such models is known to be challenging, but so is handling the stochastic nature of generative tasks. Most approaches in this domain benefit from variational methods~\cite{Kingma2014AutoEncodingVB}. These methods usually use a recurrent network~\cite{10.1162/neco.1997.9.8.1735} conditioned on a global~\cite{Babaeizadeh2018StochasticVV} or a per-frame~\cite{denton2018stochastic,babaeizadeh2021fitvid} latent variable. To model longer sequences, hierarchical variational models have been proposed ~\cite{Wichers2018HierarchicalLV,Lee2021RevisitingHA}. Another approach that scales to long sequences is keypoint-based video prediction~\cite{Minderer2019UnsupervisedLO,Kim2019UnsupervisedKL,Gao2021AccurateGK}, where they cast the problem into first keypoint dynamics prediction (with a variational method) and then pixel prediction. The usage of such methods in complex datasets with non homogeneous keypoints is yet to be seen. To overcome the blurry results that variational methods have been known for, SAVP~\cite{lee2018stochastic} combined both an adversarial loss~\cite{Goodfellow2014GenerativeAN} and a VAE~\cite{Kingma2014AutoEncodingVB}. Instead, Mathieu et al.~\cite{Mathieu2016DeepMV} used a multiscale network for video prediction. Many methods deal with motion and content separately~\cite{Sun2018ATV,Villegas2017DecomposingMA,Liang2017DualMG,Denton2017UnsupervisedLO}. The fundamental problem with using GANs for video prediction is to ensure long-term temporal consistency~\cite{Aldausari2020VideoGA}. This issue is tackled in recent works~\cite{Clark2019AdversarialVG,Luc2020TransformationbasedAV}, but at a huge computational cost.

\noindent\textbf{Transformers and Quantized Latents.} Following the success of large language models~\cite{Brown2020LanguageMA}, autoregressive transformers~\cite{NIPS2017_3f5ee243} emerged in the video synthesis domain and are replacing RNNs. However, because of the attention mechanism, transformers incur high computational cost that scales quadratically with the number of inputs. In order to scale these methods to long and higher-resolution videos, an established method is to predict vector-quantized codes~\cite{Oord2017NeuralDR} (usually obtained with VQGAN~\cite{esser2021taming}) either per frame~\cite{le2021ccvs,rakhimov2020latent,gupta2022maskvit,seo2021autoregressive} or a set of frames~\cite{yan2021videogpt}, instead of pixels.

\noindent\textbf{Diffusion Methods.} Following the impressive results of score-based diffusion models~\cite{Song2021ScoreBasedGM} on image generation~\cite{Dhariwal2021DiffusionMB,Ramesh2022HierarchicalTI}, several researchers extended these models for either video generation~\cite{ho2022video} or video prediction~\cite{voleti2022mcvd,hoppe2022diffusion,harvey2022flexible,ho2022imagen,yang2022diffusion}. Even though unconditional models can be used to approximate conditional distributions (as in video prediction)~\cite{ho2022video}, it has been shown that directly modeling the conditional distribution yields a better performance~\cite{Tashiro2021CSDICS}. MCVD~\cite{voleti2022mcvd} uses masking to train a single model capable of generating past, future, or intermediate frames. Masking allows this model to generate longer sequences by applying a moving window, even though it was only trained with a fixed number of frames. However, a common shortcoming of all these methods is that conditioning on past frames increases the number of input frames, and thus also the computational cost of training. 

Conditions for generative models are usually formulated as a fixed window of previous frames. However,  FDM~\cite{harvey2022flexible} uses a per-frame UNet~\cite{Ronneberger2015UNetCN} and attention to take a variable number of frames as input. Each input frame can be set as a conditioning input or as a prediction target, and hence this model can be conditioned on frames arbitrarily far in the past and can even predict multiple frames at the same time. More recently 3DiM~\cite{watson2022novel} introduced the idea of ``implicit'' conditioning in the task of 3D multi-view reconstruction. The idea is that at each step of the image generation in the diffusion process, the denoising network is conditioned only on a random view, instead of all the views. In this way the conditioning on multiple frames can be distributed over the denoising steps. In \methodName we extend this idea to the case of videos, where instead of views we use past frames.

Recently, \cite{lipman2022flow} introduced conditional flow matching. They showed that it generalizes diffusion models, and it achieves faster training convergence and better results than other denoising diffusion models. Thus, we adopt this framework in our generative model.There are many ways to accelerate or improve vanilla diffusion models~\cite{Kong2021OnFS,JolicoeurMartineau2021GottaGF,Dockhorn2022GENIEHD}. Among all, CCDF~\cite{Chung2022ComeCloserDiffuseFasterAC} is the most relevant to our \warm scheme. CCDF starts the backward denoising process from some time $t$ other than $T$ (which is the final time of the forward diffusion process) using an initial guess of the final output (\eg, a low-resolution sample).
In \methodName, we develop an analogous technique within the formulation of flow matching for video generation.

\begin{figure*}[t]
    \centering
    \includegraphics[width=.76\linewidth, trim=12cm 7.5cm 11cm 8cm, clip]{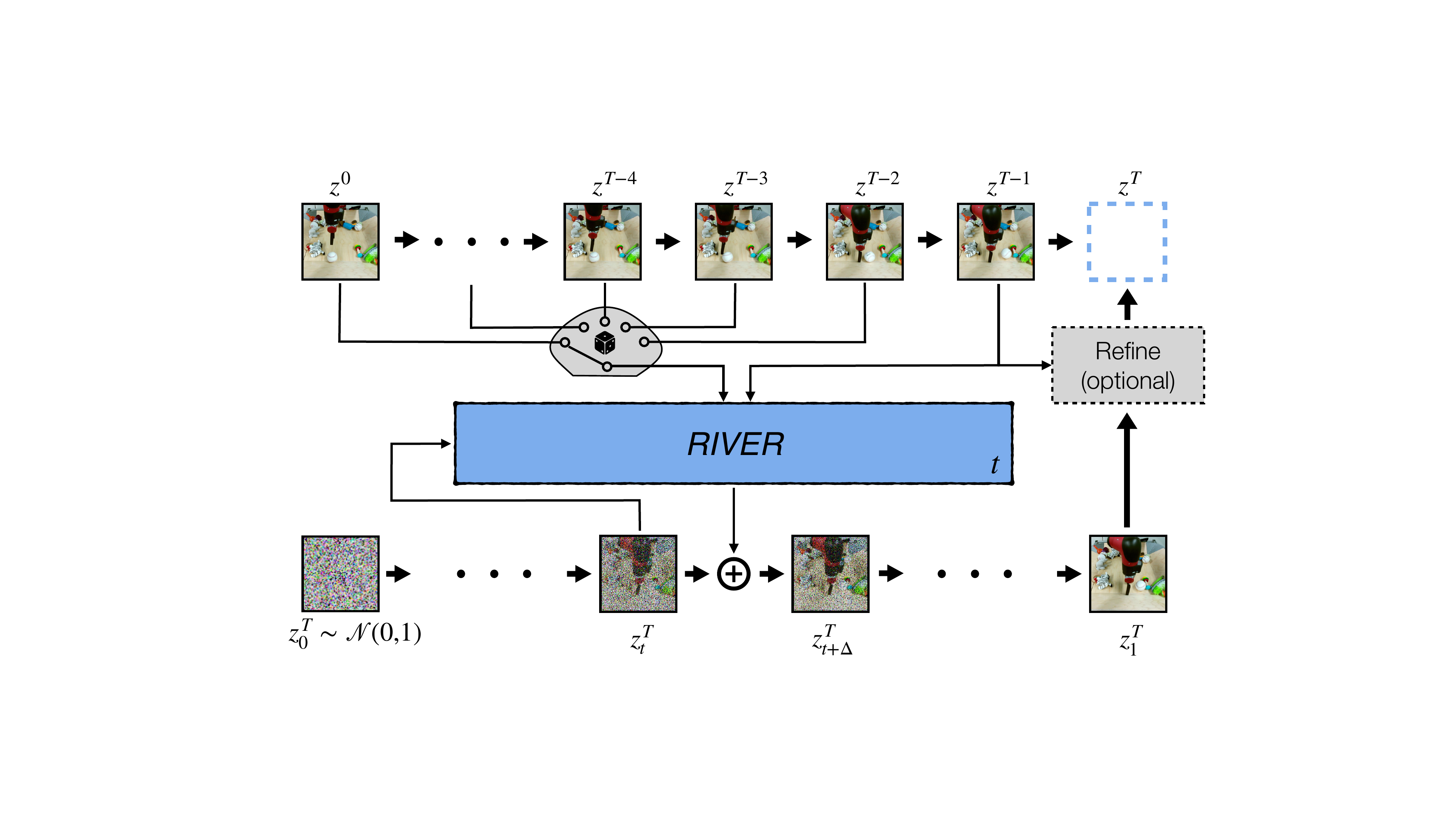}
    \caption{\aram{Inference} with \methodName. In order to generate the next frame $z^{T}$ (top-right), we sample an initial estimate from the standard normal distribution $z_t^T$ (bottom-left) and integrate the ODE~\eqref{eq:ode} by querying our model at each step with a random conditioning frame from the past $z^c$ and previous frame $z^{T-1}$ (top). \aram{We omitted the encoding/decoding for simplicity.}}\label{fig:river}
\end{figure*}

\section{Method}
\label{sec:method}

Let $\mathbf{x} = \{x^{1}, \dots, x^{m}\}$, where $x^{i} \in \mathbb{R}^{3 \times H \times W}$, be a video consisting of $m$ RGB images. The task of video prediction is to forecast the upcoming $n$ frames of a video given the first $k$ frames, where $m=n+k$. Thus, it requires modelling the following distribution:
\begin{multline}
    p(x^{k + 1}, \dots, x^{k + n} \;|\; x^{1}, \dots, x^{k}) = \\
    = \prod_{i = 1}^n p(x^{k + i} \;|\; x^{1}, \dots, x^{k + i - 1}).\label{eq:dec}
\end{multline}
The decomposition in eq.~\eqref{eq:dec} suggests an autoregressive sampling of the future frames. However, explicitly conditioning the next frame on all the past frames is computationally and memory-wise demanding. In order to overcome this issue, prior work suggests to use a recurrently updated memory variable~\cite{wang2018predrnn++,Oliu2018FoldedRN,Villegas2019HighFV,castrejon2019improved} or to restrict the conditioning window to a fixed number of frames~\cite{weissenborn2019scaling,voleti2022mcvd,hoppe2022diffusion,yang2022diffusion}. We instead propose to model each one-step predictive conditional distribution as a denoising probability density path that starts from a standard normal distribution. 
Moreover, rather than using score-based diffusion models \cite{Song2021ScoreBasedGM} to fit those paths, we choose flow matching\cite{lipman2022flow}, a simpler method to train generative models.
We further leverage the iterative nature of sampling from the learned flow and use a single random conditioning frame from the past at each iteration. This results in a simple and efficient training. An idea similar to ours was first introduced in \cite{watson2022novel} for novel view synthesis in 3D applications. In this paper, however, we made some design choices to adapt it to videos. 

\subsection{Latent Image Compression}
\label{sec:encoding}
Although we could operate directly on the pixels of the frames $x^i$, we introduce a compression step that reduces the dimensionality of the data samples and thus the overall numerical complexity of our approach.
Given a dataset of videos $D$, we train a VQGAN \cite{esser2021taming} on single frames from that dataset. The VQGAN consists of an encoder ${\cal E}$ and a decoder ${\cal D}$ and allows to learn a perceptually rich latent codebook through a vector quantization bottleneck and an adversarial reconstruction loss~\cite{Oord2017NeuralDR}. A trained VQGAN is then used to compress the images to much lower resolution feature maps. That is, $z = {\cal E}(x) \in \mathbb{R}^{c\times \frac{H}{f} \times \frac{W}{f}}$, where $x \in \mathbb{R}^{3\times H \times W}$. Commonly used values for $c$ are 4 or 8 and for $f$ are 8 or 16, which means that a $256 \times 256$ image can be downsampled to up to a $16 \times 16$ grid. Following \cite{rombach2022high}, we let the decoder ${\cal D}$ absorb the quantization layer and work in the pre-quantized latent space. Further in the paper, when referring to video frames we always assume that they are encoded in the latent space of a pretrained VQGAN.

\subsection{Flow Matching}
\label{sec:flow}

Flow matching was introduced in \cite{lipman2022flow} as a simpler albeit more general and more efficient alternative to diffusion models~\cite{Ho2020DenoisingDP}.  \aram{A similar framework incorporating straight flows has also been proposed independently in \cite{liu2022flow, albergo2022building}.} We assume that we are given samples from an unknown data distribution $q(z)$. In our case, the data sample $z$ is the encoding of a video frame $x$ via VQGAN. 
The aim of flow matching is to learn a temporal vector field $v_t(z): [0, 1] \times \mathbb{R}^d \rightarrow \mathbb{R}^d$, with $t\in [0,1]$, such that the following ordinary differential equation (ODE) 
\begin{align}\label{eq:ode}
    \dot \phi_t(z) &= v_t(\phi_t(z)) \\
    \phi_0(z) &= z
\end{align}
defines a flow $\phi_t(z): [0, 1] \times \mathbb{R}^d \rightarrow \mathbb{R}^d$ that pushes $p_0(z) = {\cal N}(z \,|\, 0, 1)$ towards some distribution $p_1(z) \approx q(z)$ along some probability density path $p_t(z)$. That is,
    $p_t = [\phi_t]_* p_0$,
where $[\cdot]_*$ denotes the push-forward operation.
If one were given a predefined probability density path $p_t(z)$ and the corresponding vector field $u_t(z)$, then one could parameterize $v_t(z)$ with a neural network and solve
\begin{align}\label{eq:fm}
    \min_{v_t} \mathbb{E}_{t, p_t(z)} \| v_t(z) - u_t(z) \|^2.
\end{align}
However, this would be unfeasible in the general case, because typically we do not have access to $u_t(z)$. Lipman \etal~\cite{lipman2022flow} suggest to instead define a conditional flow $p_t(z \,|\, z_1)$ and the corresponding conditional vector field $u_t(z \,|\, z_1)$ per sample $z_1$ in the dataset and solve
\begin{align}
    \min_{v_t} \mathbb{E}_{t, p_t(z \,|\, z_1), q(z_1)} \| v_t(z) - u_t(z \,|\, z_1) \|^2.
    \label{eq:simplified}
\end{align}
This formulation enjoys two remarkable properties: 1) all the quantities can be defined explicitly; 2) Lipman \etal~\cite{lipman2022flow} show that solving eq.~\eqref{eq:simplified} is guaranteed to converge to the same result as in eq.~\eqref{eq:fm}.
The conditional flow can be explicitly defined such that all intermediate distributions are Gaussian. Moreover, Lipman \etal~\cite{lipman2022flow} show that a linear transformation of the Gaussians' parameters yields the best results in terms of convergence and sample quality. They define $p_t(z \,|\, z_1) = {\cal N}(z \,|\, \mu_t(z_1), \sigma^2_t(z_1))$, with 
$\mu_t(x) = tx_1$ and $\sigma_t(x) = 1- (1-\sigma_\text{min})t$. With these choices, the corresponding target vector field is given by
\begin{align}\label{eq:u}
    u_t(z \,|\, z_1) = \frac{z_1 - (1 - \sigma_{\text{min}}) z}{1 - (1 - \sigma_{\text{min}}) t}.
\end{align}
Sampling from the learned model can be obtained by first sampling $z_0 \sim {\cal N}(z \,|\, 0, 1)$ and then numerically solving eq.~\eqref{eq:ode} for $z_1 = \phi_1(z_0)$. 

\subsection{Video Prediction}
\label{sec:vidpred}
We introduce the main steps to train and use \methodName. First, as described in sec.~\ref{sec:encoding} we use a per-frame perceptual autoencoder to reduce the dimensionality of data. Since the encoding is per-frame and thus the reconstruction error could be temporally inconsistent, we improve the quality of a generated video by also introducing an optional small autoregressive refinement step in the decoding network. Second, we train a denoising model via flow matching in the space of encoded frames with our distributed conditioning. Moreover, we accelerate the video generation by introducing a \warm procedure.

\begin{algorithm}[t]
    \caption{Video Flow Matching with \methodName}\label{alg:training}
    \begin{algorithmic}
        \STATE Input: dataset of videos $D$, number of iterations $N$
        \FOR{$i$ in range(1, $N$)}
        \STATE Sample a video $\mathbf{x}$ from the dataset $D$
        \STATE Encode it with a pre-trained VQGAN to obtain $z$
        \STATE Choose a random target frame $z^{\tau}, \tau \in \{3, \dots |\mathbf{x}|\}$
        \STATE Sample a timestamp $t \sim U[0, 1]$
        \STATE Sample a noisy observation $z \sim p_t(z \,|\, z^{\tau})$
        \STATE Calculate $u_t(z \,|\, z^{\tau})$
        \STATE Sample a condition frame $z^{c}, c \in \{1, \dots \tau - 2\}$
        \STATE Update the parameters $\theta$ of $v_t$ via gradient descent 
        \begin{align}
            \nabla_{\theta} \| v_{t}(z \,|\, z^{\tau-1}, z^{c}, \tau - c \,; \theta) - u_t(z \,|\, z^{\tau}) \|^2
        \end{align}
        \ENDFOR
    \end{algorithmic}
\end{algorithm}

\noindent\textbf{Training.}
We adapt Flow Matching~\cite{lipman2022flow} to video prediction by letting the learned vector field $v_t$ condition on the past context frames. Furthermore, 
we randomize the conditioning at each denoising step 
to only 2 frames. This results in a very simple training procedure, which is described in Algorithm~\ref{alg:training}. Given a training video $\mathbf{z} = \{z^{1}, \dots, z^{m}\}$ (pre-encoded with VQGAN), we randomly sample a target frame $z^{\tau}$ and a random (diffusion) timestep $t \sim U[0, 1]$. We can then draw a sample from the conditional probability distribution $z \sim p_t(z \,|\, z^{\tau})$ and calculate the target vector field $u_t(z \,|\, z^{\tau})$ using eq.~\eqref{eq:u}. We then sample another index $c$ uniformly from $\{1, \dots, \tau - 2\}$ and use $z^{c}$, which we call \emph{context frame}, together with $z^{\tau-1}$, which we call \emph{reference frame}, as the two conditioning frames. 
Later, we show that the use of the reference is crucial for the network to learn the scene motion, since one context frame carries very little information about such motion. The vector field regressor $v_t$ is then trained to minimize the following objective
\begin{align}
    {\cal L}_{\text{FM}}(\theta) = \| v_t(z \,|\, z^{\tau-1}, z^{c}, \tau - c \,; \theta) - u_t(z \,|\, z^{\tau}) \|^2,
\end{align}
where $\theta$ are the parameters of the model.
Note that at no point during the training the whole video sequence must be stored or processed. Moreover, the generation of frames is never needed, which further simplifies the training process.

\noindent\textbf{Inference.}
At inference time, in order to generate the $T$-th frame, we start from sampling an initial estimate $z_0^T$ from the standard normal distribution (see Figure~\ref{fig:river}). We then use an ODE solver to integrate the learned vector field along the time interval $[0, 1]$. At each integration step, the ODE solver queries the network for $v_t(z_t^T \,|\, z^{T - 1}, z^c, T - c)$, where $c \sim U\{1, \dots, T - 2\}$. In the simplest case, the Euler step of the ODE integration takes the form
\begin{align}
    z_{t_{i+1}}^T = z_{t_{i}}^T + \frac{1}{N} v_{t_{i}}(z_{t_{i}}^T \,|\, z^{T - 1}, z^{c_i}, T - {c_i}), \\
    \quad c_i \sim U\{1, \dots, T - 2\}, \\
    z_{t_0}^T \sim {\cal N}(z \,|\, 0, 1), \\
    t_i = \frac{i}{N}, \; i \in \{0, \dots, N - 1\},
\end{align}
where $N$ is the number of integration steps.
We then use $z_1^T$ as an estimate of $z^{T}$. 


\noindent\textbf{Refinement.}
\sepehr{When using a per-frame VQGAN~\cite{esser2021taming}, the autoencoded videos may not always be temporally consistent. To address this issue without incurring a significant computational cost, we optionally utilize a refinement network that operates in the pixel space. This deep convolutional network, based on the architecture of RCAN~\cite{Zhang2018ImageSU}, is trained using the previous frame and the decoded next frame to refine the second frame. 
We train the model using an $L_2$ and a perceptual loss by refining 16 consecutive frames independently and then by feeding all frames to a perceptual network (I3D~\cite{Carreira2017QuoVA} in our case). We train the refinement network separately after training the autoencoder.}

\begin{figure}[t]
    \centering
    \includegraphics[width=0.9\linewidth,trim=0 0 1cm 1.5cm, clip]{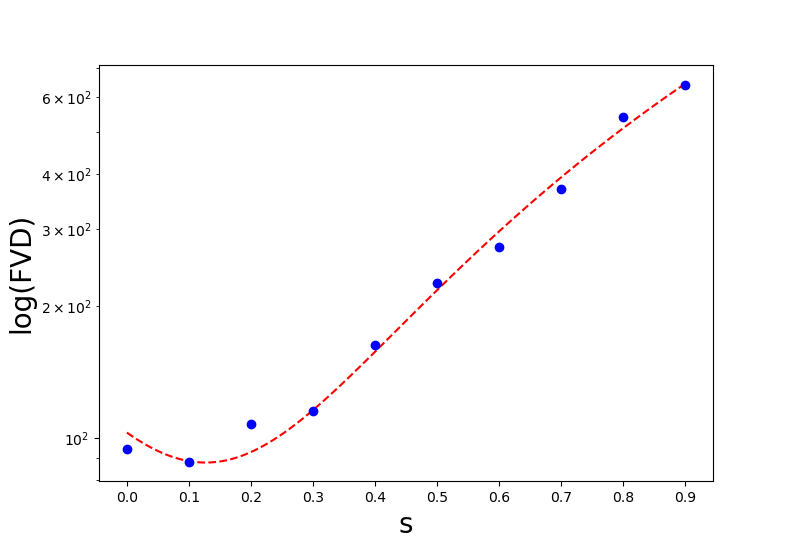}
    \caption{Higher values of $s$ for \warm lead to faster sampling, but worse FVD. Interestingly, $s=0.1$ acts like the truncation trick~\cite{Marchesi2017MegapixelSI,Brock2019LargeSG} and slightly improves the FVD.}
    \label{fig:s_tradeoff}
\end{figure}

\noindent\textbf{Sampling Speed.}
A common issue of models based on denoising processes is the sampling speed, as the same denoising network is queried multiple times along the denoising path in order to generate an image. This is even more apparent for the video domain, where the generation speed scales with the number of frames to generate. Some video diffusion models~\cite{harvey2022flexible,voleti2022mcvd} overcome this issue by sampling multiple frames at a time. However, the price they have to pay is the inability to generate arbitrarily long videos. We instead leverage the temporal smoothness of videos, that is, the fact that subsequent frames in a video do not differ much. This allows us to use a noisy previous frame as the initial condition of the ODE instead of pure noise. More precisely, instead of starting the integration from $z_0 \sim {\cal N}(z \,|\, 0, 1)$, we start at $z_s' \sim p_s(z \,|\, z^{T - 1})$, where $1 - s$ is the speed up factor. We call this technique \emph{\warm}. 
Intuitively, larger $s$ results in a lower variability in the future frames. Moreover, we found that starting closer to the end of the integration path reduces the magnitude of the motion in the generated videos, since the model is required to sample closer to the previous frame. Therefore, there is a tradeoff between the sampling speed and the quality of the samples. We further emphasize this tradeoff by computing the FVD~\cite{Unterthiner2018TowardsAG} of the generated videos depending on the speed up factor $1 - s$ (see Figure~\ref{fig:s_tradeoff}).

\begin{figure*}[t]
    \centering
    \hspace{-.3cm}
    \begin{tabular}{@{}r@{}c@{\hspace{1mm}}c@{\hspace{0.5mm}}c@{\hspace{0.5mm}}c@{\hspace{0.5mm}}c@{\hspace{0.5mm}}c@{\hspace{0.5mm}}c@{\hspace{0.5mm}}c@{\hspace{0.5mm}}c@{}}
    & \makebox[1.85cm][c]{\small last context frame} & \multicolumn{5}{c}{\small time $\rightarrow$}\\
    \rotatebox{90}{\makebox[1.85cm][c]{\small GT}} &
    \multicolumn{1}{c@{\hspace{0.5mm}}?@{\hspace{0.5mm}}}{\includegraphics[width=1.85cm]{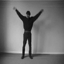}} & \includegraphics[width=1.85cm]{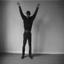} & 
    \includegraphics[width=1.85cm]{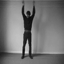} & 
    \includegraphics[width=1.85cm]{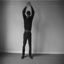} & 
    \includegraphics[width=1.85cm]{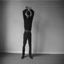} & 
    \includegraphics[width=1.85cm]{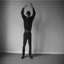} &
    \includegraphics[width=1.85cm]{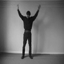} & 
    \includegraphics[width=1.85cm]{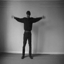} & 
    \includegraphics[width=1.85cm]{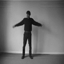} \\
    & \multicolumn{1}{r@{\hspace{0.5mm}}?@{\hspace{0.5mm}}}{\rotatebox{90}{\makebox[1.85cm][c]{predicted}}} &
    \includegraphics[width=1.85cm]{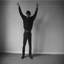} &
    \includegraphics[width=1.85cm]{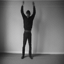} &
    \includegraphics[width=1.85cm]{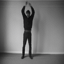} & 
    \includegraphics[width=1.85cm]{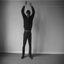} & 
    \includegraphics[width=1.85cm]{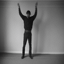} &
    \includegraphics[width=1.85cm]{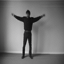} & 
    \includegraphics[width=1.85cm]{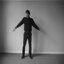} & 
    \includegraphics[width=1.85cm]{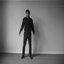} \\
    \rotatebox{90}{\makebox[1.85cm][c]{\small GT}} & \multicolumn{1}{c@{\hspace{0.5mm}}?@{\hspace{0.5mm}}}{\includegraphics[width=1.85cm]{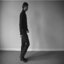}} & 
    \includegraphics[width=1.85cm]{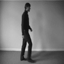} & 
    \includegraphics[width=1.85cm]{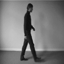} & 
    \includegraphics[width=1.85cm]{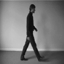} & 
    \includegraphics[width=1.85cm]{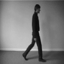} & 
    \includegraphics[width=1.85cm]{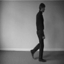} &
    \includegraphics[width=1.85cm]{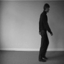} & 
    \includegraphics[width=1.85cm]{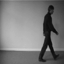} & 
    \includegraphics[width=1.85cm]{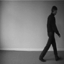} \\
    & \multicolumn{1}{r@{\hspace{0.5mm}}?@{\hspace{0.5mm}}}{\rotatebox{90}{\makebox[1.85cm][c]{predicted}}} &
    \includegraphics[width=1.85cm]{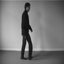} & 
    \includegraphics[width=1.85cm]{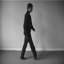} & 
    \includegraphics[width=1.85cm]{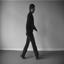} & 
    \includegraphics[width=1.85cm]{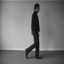} & 
    \includegraphics[width=1.85cm]{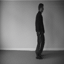} &
    \includegraphics[width=1.85cm]{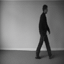} & 
    \includegraphics[width=1.85cm]{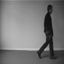} & 
    \includegraphics[width=1.85cm]{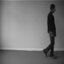}
    \end{tabular}
    \caption{Video prediction on the \emph{KTH} dataset. In order to predict the future frames, the model conditions on the first 10 (context) frames. Of this sequence, only the last context frame is shown. \sepehr{By definition, a proper stochastic predictive model generates realistic predictions of future frames that do not necessarily match the GT data.}  
    }
    \label{fig:kth}
\end{figure*}

\begin{table}[t]
    \centering
    \begin{tabular*}{\linewidth}{l@{\extracolsep{\fill}}ccc@{}}
    \toprule
        Method & FVD$\downarrow$ & PSNR$\uparrow$ & SSIM$\uparrow$\\
    \hline
        \multicolumn{4}{@{}l}{\textit{10$\rightarrow$30}}\\
        SRVP~\cite{Franceschi2020StochasticLR} & 222 & 29.7 & \textbf{0.87} \\
        SLAMP~\cite{Akan2021SLAMPSL} & 228 & 29.4 & \textbf{0.87} \\
        MCVD~\cite{voleti2022mcvd} & 323 & 27.5 & 0.84 \\
        \methodName (ours) & \textbf{180} & \textbf{30.4} & \aram{0.86} \\
    \hline
    \multicolumn{4}{@{}l}{\textit{10$\rightarrow$40}}\\
        MCVD~\cite{voleti2022mcvd} & 276.7 & 26.4 & 0.81 \\
        GridKeypoints~\cite{gao2021accurate} & \textbf{144.2} & 27.1 & \textbf{0.84} \\
        \methodName (ours) & 170.5 & \textbf{29.0} & \aram{0.82} \\
    \bottomrule
    \end{tabular*}
    \caption{\emph{KTH} dataset evaluation. The evaluation protocol is to predict the next 30/40 frames given the first 10 frames.}
    \label{tab:kth}
\end{table}

\subsection{Implementation}
\label{sec:impl}

A commonly leveraged architecture for flow matching and diffusion models is UNet~\cite{Ronneberger2015UNetCN}. However, we found that training UNet could be time demanding. Instead, we propose to model $v_t(z \,|\, z^{\tau - 1}, z^{c}, \tau - c \,; \theta)$ with the recently introduced U-ViT~\cite{bao2022all}. U-ViT follows the standard ViT~\cite{Dosovitskiy2021AnII} architecture and adds several long skip-connections, like in UNet. This design choice allows U-ViT to achieve on par or better results than UNet on image generation benchmarks with score-based diffusion models. 

The inputs to the network are $\sfrac{HW}{f^2}$ tokens constructed by concatenating $z, z^{\tau - 1}$ and $z^{c}$ in feature axis as well as one additional time embedding token $t$ that makes the network time-dependent. We additionally add spatial position encondings to the image tokens and augment $z^{\tau - 1}$ and $z^{c}$ with an encoded relative distance $\tau - c$ to let the network know how far in the past the condition is. That is, the overall input to the network is of size $[\sfrac{HW}{f^2} + 1, 3 \times d]$, where the first dimension refers to the number of tokens, while the second refers to the number of channels. 
For further details, see the supplementary material.
 
\section{Experiments}
\label{sec:exp}

In section~\ref{sec:cond}, we report our results on several video prediction benchmarks. We evaluate our method using standard metrics, such as FVD~\cite{Unterthiner2018TowardsAG}, PSNR and SSIM~\cite{Wang2004ImageQA}. 
We additionally show in section~\ref{sec:planning} that our model is able to perform visual planning. Video generation is demonstrated in section~\ref{sec:generation}. 
Note that if not explicitly specified, we use the model without the refinement stage and with $s = 0$ in \warm. For additional results and training details, see the supplementary material.

\subsection{Conditional Video Prediction}
\label{sec:cond}

We test our method on \if\kitti1 3 \else 2 \fi datasets. 
First, to assess the ability of \methodName to generate structured human motion, we test it on the \textbf{KTH} dataset~\cite{schuldt2004recognizing}. KTH is a dataset containing 6 different human actions performed by 25 subjects in different scenarios. We follow the standard evaluation protocol predicting 30/40 future frames conditioned on the first 10 at a $64\times 64$ pixel resolution. The results are reported in Table~\ref{tab:kth}. We show that \methodName achieves state of the art prediction quality compared to prior methods that do not use domain-specific help. For instance, \cite{gao2021accurate} models the motion of the keypoints, which works well for human-centric data, but does not apply to general video generation. Figure~\ref{fig:kth} shows qualitative results.

\if\kitti1
We also test on \textbf{KITTI}~\cite{Geiger2012AreWR} in order to assess our model's ability to work with dynamic backgrounds. KITTI is a video dataset captured by a setup mounted on a car navigating streets in Berlin. The scenes in the dataset are partially observable, so the model has to hallucinate in order to predict the future, which makes this test case extremely challenging. We follow the preprocessing and testing format proposed in~\cite{} and~\cite{}, which is to predict 25 future frames given first 5. The small size of the dataset (the training set contains only 57 videos) further complicates the task, as the models tend to overfit. In order to reduce this issue, during the training we apply the same augmentations as in~\cite{babaeizadeh2021fitvid}. The results are reported in Table~\ref{tab:kitti}.

\begin{table}[t]
    \centering
    \begin{tabular*}{\linewidth}{l@{\extracolsep{\fill}}cccc@{}}
    \toprule
        Method & FVD$\downarrow$ & PSNR$\uparrow$ & SSIM$\uparrow$ & LPIPS$\downarrow$\\
    \hline
        HARP\cite{seo2021autoregressive} & 482.9 & - & - & 0.191\\
        GHVAE~\cite{wu2021greedy} & 552.9 & 15.8 & 0.51 & 0.286\\
        FitVid~\cite{babaeizadeh2021fitvid} & 884.5 & 17.1 & 0.49 & 0.217\\
        Transframer~\cite{nash2022transframer} & 260.4 & 17.9 & 0.54 & 0.112\\
    \hline
        \methodName \\ (1 view, 100 steps) & 1013 & 13.5 & 0.87 & 0.33\\
    \bottomrule
    \end{tabular*}
    \caption{\emph{KITTI} dataset evaluation. The evaluation protocol is to predict 25 future frames given initial 5.}
    \label{tab:kitti}
\end{table}
\fi

\begin{figure*}[t]
    \centering
    \begin{tabular}{@{}c@{\hspace{1mm}}c@{\hspace{0.5mm}}c@{\hspace{0.5mm}}c@{\hspace{0.5mm}}c@{}}
    \makebox[1.7cm][c]{initial frame} & \multicolumn{4}{c}{time $\rightarrow$}\\
        
    \multicolumn{1}{c@{\hspace{0.5mm}}?@{\hspace{0.5mm}}}{\hspace{-.3cm}\includegraphics[width=3.45cm]{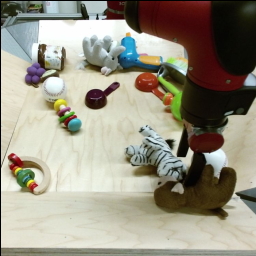}} & 
    \includegraphics[width=3.45cm]{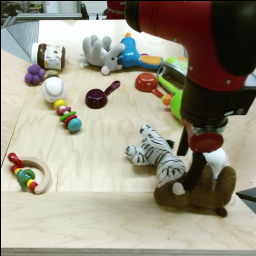} &
    \includegraphics[width=3.45cm]{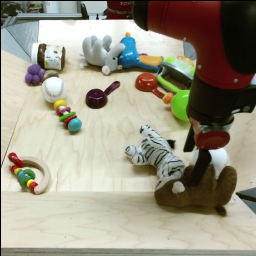} &
    \includegraphics[width=3.45cm]{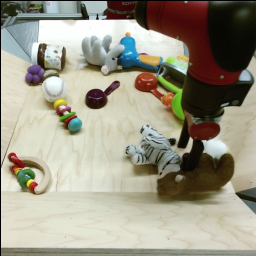} &
    \includegraphics[width=3.45cm]{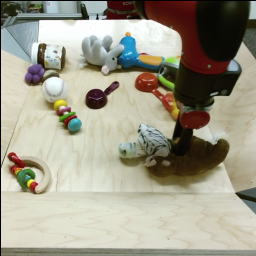}\\
    \end{tabular}
    \caption{Video prediction on the \emph{BAIR} dataset. The model predicts future frames conditioned on a single initial frame. Thanks to VQGAN, \methodName can be used to generate high resolution videos. 
    }
    \label{fig:bair}
\end{figure*}

Additionally, in Table~\ref{tab:bair} we evaluate the capability of \methodName to model complex interactions on \noindent\textbf{BAIR}~\cite{Ebert2017SelfSupervisedVP}, which is a dataset containing around 44K clips of a robot arm pushing toys on a flat square table. For BAIR, we generate and refine 15 future frames conditioned on one initial frame at a 64$\times$64 pixel resolution. Due to the high stochasticity of motion in the BAIR dataset, the standard evaluation protocol in~\cite{babaeizadeh2021fitvid} is to calculate the metrics by comparing 100$\times$256 samples to 256 random test videos (\ie, 100 generated videos for each test video, by starting from the same initial frame as the test example). \aram{Additionally, we report the compute (memory in GB and hours) needed to train the models. \methodName reaches a tradeoff between the FVD and the compute and generates smooth realistic videos while requiring much less computational effort (see also Figure~\ref{fig:tradeoff}). In addition, we calculate the FVD vs the autoencoded test set, as we find that FVD (like FID~\cite{Parmar2022OnAR}) can be affected even by different interpolation techniques. This way we eliminate the influence of potential autoencoding artifacts on the metrics in order to assess the consistency of the motion only. In fact, there is an improvement of about 30\% in the FVD.} 
Furthermore, although the standard benchmark on BAIR uses 64$\times$64 pixels resolution, with the help of the perceptual compression, we are able to generate higher-resolution videos under the same training costs. See Figure~\ref{fig:bair} for qualitative results on the \emph{BAIR} dataset at 256$\times$256 resolution. Finally, we would like to point out that we observed DDPM fail to converge on BAIR, which further justifies our choice of flow matching (see also the appendix).

\begin{table}[t]
    \centering
    \begin{tabularx}{\linewidth}{@{}l@{\extracolsep{\fill}}rcc@{}}
    \toprule
        Method & FVD$\downarrow$ & Mem (GB) & Hours\\
    \hline
        TriVD-GAN-FP~\cite{luc2020transformation} & 103.0 & 1024 & 280 \\
        Video Transformer~\cite{weissenborn2019scaling} (L) & 94.0 & 512 & 336 \\
        CCVS~\cite{le2021ccvs} (low res) & 99.0 & 128 & 40 \\
        CCVS~\cite{le2021ccvs} (high res) & 80.0 & - & - \\
        LVT~\cite{rakhimov2020latent} ($n_c=4$) & 125.8 & 128 & 48 \\
        FitVid~\cite{babaeizadeh2021fitvid} & 93.6 & 1024 & 288 \\
        MaskViT~\cite{gupta2022maskvit} & 93.7 & - & - \\
        MCVD~\cite{voleti2022mcvd} (concat) & 98.8 & 77 & 78 \\
        MCVD~\cite{voleti2022mcvd} (spatin) & 103.8 & 86 & 50 \\
        NÜWA~\cite{Wu2022NWAVS} & 86.9 & 2560 & 336 \\
        RaMViD~\cite{hoppe2022diffusion} & 84.2 & 320 & 72 \\
        VDM~\cite{ho2022video}	& 66.9 & - & - \\
    \hline
        \methodName \textit{w/ refine} & 106.1 & 25 &  25 \\
        \methodName \textit{w/o refine} & 145.8 & - & - \\
        \methodName \textit{w/o refine vs ae GT} & 73.5 & - & - \\
    \bottomrule
    \end{tabularx}
    \caption{\emph{BAIR} dataset evaluation. We follow the standard evaluation protocol, which is to predict 15 future frames given 1 initial frame. The common way to compute the FVD is to compare 100$\times$256 generated sequences to 256 randomly sampled test videos. 
    \aram{Additionally, we report the numbers of the network without the refinement stage versus the original ground truth (RIVER \textit{w/o refine}) and the autoencoded ground truth (RIVER \textit{w/o refine vs ae GT}) to highlight the influence of the VQGAN's artifacts on the assessment of the motion consistency.}}
    \label{tab:bair}
\end{table}


\subsection{Visual Planning}
\label{sec:planning}

One way to show the ability of the model to learn the dynamics of the environment is to do planning~\cite{finn2017deep, finn2016unsupervised, xie2021learning}.
With a small change to the training of our model, \methodName is able to infill the video frames given the source and the target images. The only change to be done to the model is to remove the reference frame and to let two condition frames be sampled from both the future frames and the past ones.
At inference time, given the source and the target frames, our model sequentially infills the frames between those. We show in Figure~\ref{fig:planning} some qualitative results of video interpolation on the CLEVRER~\cite{Yi2020CLEVRERCE} dataset, which is a dataset containing 10K training clips capturing a synthetic scene with multiple objects interacting with each other through collisions. It is a dataset suitable for planning, as it allows to show the ability of the method to model the dynamics of the separate objects and their interactions. We test our model at the 128$\times$128 pixels resolution. Note how the model has learned the interactions between the objects and is able to manipulate the objects in order to achieve the given goals.

\begin{figure*}[t]
    \centering
    \begin{tabular}{@{}c@{\hspace{0.mm}}c@{\hspace{0.mm}}c@{\hspace{0.mm}}c@{\hspace{0.mm}}c@{}}
    \makebox[2.5cm][c]{source frame} & \multicolumn{3}{c}{time $\rightarrow$} & \makebox[2.5cm][c]{target frame}\\
    \tcbox[hbox, size=fbox, graphics options={width=2.5cm}, colback=green!20,]{\includegraphics[width=2.5cm]{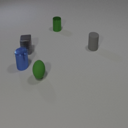}} & 
    \tcbox[hbox, size=fbox, graphics options={width=2.5cm}, colframe=white, colback=white!30,]{\includegraphics[width=2.5cm]{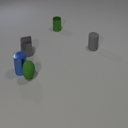}} &
    \tcbox[hbox, size=fbox, graphics options={width=2.5cm}, colframe=white, colback=white!30,]{\includegraphics[width=2.5cm]{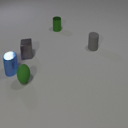}} &
    \tcbox[hbox, size=fbox, graphics options={width=2.5cm}, colframe=white, colback=white!30,]{\includegraphics[width=2.5cm]{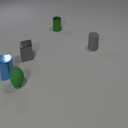}} &
    \tcbox[hbox, size=fbox, graphics options={width=2.5cm}, colback=green!20,]{\includegraphics[width=2.5cm]{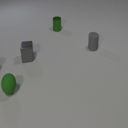}} \\
    \tcbox[hbox, size=fbox, graphics options={width=2.5cm}, colback=green!20,]{\includegraphics[width=2.5cm]{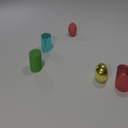}} & 
    \tcbox[hbox, size=fbox, graphics options={width=2.5cm}, colframe=white, colback=white!30,]{\includegraphics[width=2.5cm]{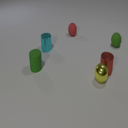}} &
    \tcbox[hbox, size=fbox, graphics options={width=2.5cm}, colframe=white, colback=white!30,]{\includegraphics[width=2.5cm]{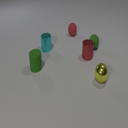}} &
    \tcbox[hbox, size=fbox, graphics options={width=2.5cm}, colframe=white, colback=white!30,]{\includegraphics[width=2.5cm]{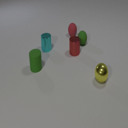}} &
    \tcbox[hbox, size=fbox, graphics options={width=2.5cm}, colback=green!20,]{\includegraphics[width=2.5cm]{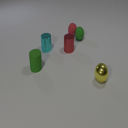}} \\
    \end{tabular}
    \caption{Visual planning with \methodName on the \emph{CLEVRER} dataset. Given the source and the target frames, \methodName infills the frames inbetween. Note how the model manipulates the objects by forcing them to interact in order to achieve the goal. In some cases this even requires introducing new objects into the scene. 
    }
    \label{fig:planning}
\end{figure*}


\begin{figure*}[t]
    \centering
    \begin{tabular}{@{}c@{\hspace{0.5mm}}c@{\hspace{0.5mm}}c@{\hspace{0.5mm}}c@{\hspace{0.5mm}}c@{\hspace{0.5mm}}c@{\hspace{0.5mm}}c@{\hspace{0.5mm}}c@{\hspace{0.5mm}}c@{}}
        \multicolumn{9}{c}{time $\rightarrow$} \\ 
        \includegraphics[width=1.85cm]{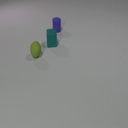} &
        \includegraphics[width=1.85cm]{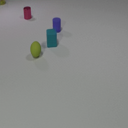} &
        \includegraphics[width=1.85cm]{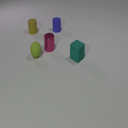} &
        \includegraphics[width=1.85cm]{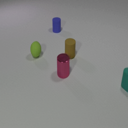} &
        \includegraphics[width=1.85cm]{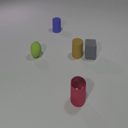} &
        \includegraphics[width=1.85cm]{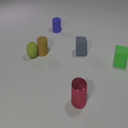} &
        \includegraphics[width=1.85cm]{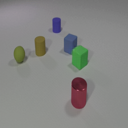} &
        \includegraphics[width=1.85cm]{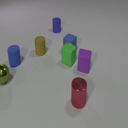} &
        \includegraphics[width=1.85cm]{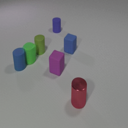} \\
        \includegraphics[width=1.85cm]{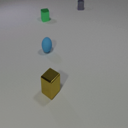} &
        \includegraphics[width=1.85cm]{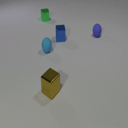} &
        \includegraphics[width=1.85cm]{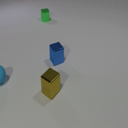} &
        \includegraphics[width=1.85cm]{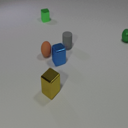} &
        \includegraphics[width=1.85cm]{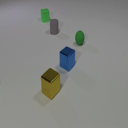} &
        \includegraphics[width=1.85cm]{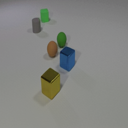} &
        \includegraphics[width=1.85cm]{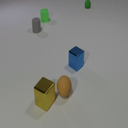} &
        \includegraphics[width=1.85cm]{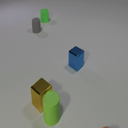} &
        \includegraphics[width=1.85cm]{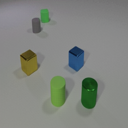} \\
        \includegraphics[width=1.85cm]{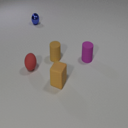} &
        \includegraphics[width=1.85cm]{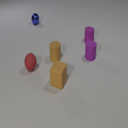} &
        \includegraphics[width=1.85cm]{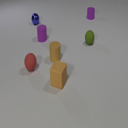} &
        \includegraphics[width=1.85cm]{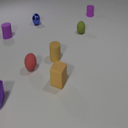} &
        \includegraphics[width=1.85cm]{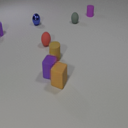} &
        \includegraphics[width=1.85cm]{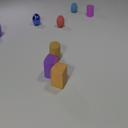} &
        \includegraphics[width=1.85cm]{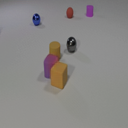} &
        \includegraphics[width=1.85cm]{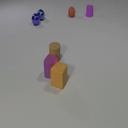} &
        \includegraphics[width=1.85cm]{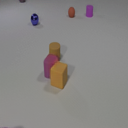} \\
        $T=1$ & $T=15$ & $T=30$ & $T=45$ & $T=60$ & $T=75$ & $T=90$ & $T=105$ & $T=120$ \\
    \end{tabular}
    \caption{\sepehr{Long video generation examples on the \emph{CLEVRER} dataset. We generate the first frame and predict the next frames.}}
    \label{fig:generation}
\end{figure*}

\subsection{Video Generation}
\label{sec:generation}

\sepehr{\methodName can be easily adapted to support \emph{video generation}. Inspired by the classifier-free guidance~\cite{ho2022classifier} we train a single model to both generate (the first frame of a video) and predict the next frames by simply feeding noise instead of the condition frames 10\% of the times during training. Then, during inference we generate the first frame and then predict the rest of the video given the first frame. Figure~\ref{fig:generation} shows our results for video generation on CLEVRER~\cite{Yi2020CLEVRERCE} ($\text{FVD}=23.63$). Other methods~\cite{Mei2022VIDMVI,Yu2022GeneratingVW,Skorokhodov2021StyleGANVAC} have difficulties in modeling the motions and interactions of objects. For videos and qualitative comparisons, visit our website\footnote{\url{https://araachie.github.io/river}}.}

\begin{figure*}[t]
    \centering
    \hspace{-.3cm}
    \begin{tabular}{@{}r@{}c@{\hspace{1mm}}c@{\hspace{0.5mm}}c@{\hspace{0.5mm}}c@{\hspace{0.5mm}}c@{\hspace{0.5mm}}c@{\hspace{0.5mm}}c@{\hspace{0.5mm}}c@{\hspace{0.5mm}}c@{}}
    {} & \makebox[1.7cm][c]{last context frame} & \multicolumn{8}{c}{time $\rightarrow$}\\
    \rotatebox{90}{\makebox[1.85cm][c]{GT}} &
    \multicolumn{1}{c@{\hspace{0.5mm}}?@{\hspace{0.5mm}}}{\includegraphics[width=1.85cm]{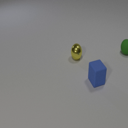}} &
    \includegraphics[width=1.85cm]{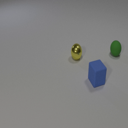} & 
    \includegraphics[width=1.85cm]{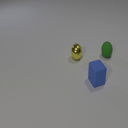} & 
    \includegraphics[width=1.85cm]{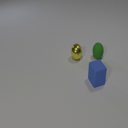} & 
    \includegraphics[width=1.85cm]{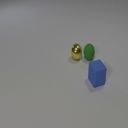} & 
    \includegraphics[width=1.85cm]{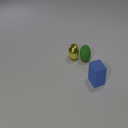} &
    \includegraphics[width=1.85cm]{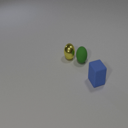} & 
    \includegraphics[width=1.85cm]{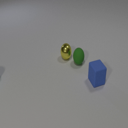} & 
    \includegraphics[width=1.85cm]{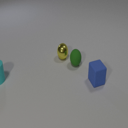} \\
    & \multicolumn{1}{c@{\hspace{0.5mm}}?@{\hspace{0.5mm}}}{\rotatebox{45}{\makebox[1.85cm][c]{w/ reference}}} & 
    \includegraphics[width=1.85cm]{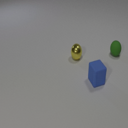} &
    \includegraphics[width=1.85cm]{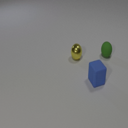} &
    \includegraphics[width=1.85cm]{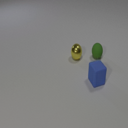} &
    \includegraphics[width=1.85cm]{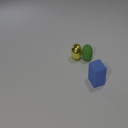} &
    \includegraphics[width=1.85cm]{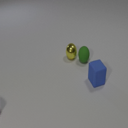} &
    \includegraphics[width=1.85cm]{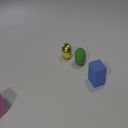} &
    \includegraphics[width=1.85cm]{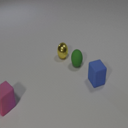} &
    \includegraphics[width=1.85cm]{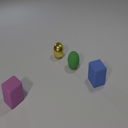} \\
    & \multicolumn{1}{c@{\hspace{0.5mm}}?@{\hspace{0.5mm}}}{\rotatebox{45}{\makebox[1.85cm][c]{w/o reference}}} & 
    \includegraphics[width=1.85cm]{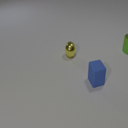} &
    \includegraphics[width=1.85cm]{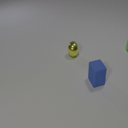} &
    \includegraphics[width=1.85cm]{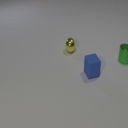} &
    \includegraphics[width=1.85cm]{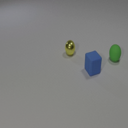} &
    \includegraphics[width=1.85cm]{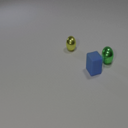} &
    \includegraphics[width=1.85cm]{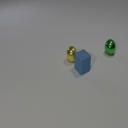} &
    \includegraphics[width=1.85cm]{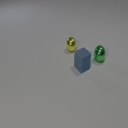} &
    \includegraphics[width=1.85cm]{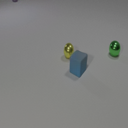} \\
    \end{tabular}
    \caption{Video prediction on the \emph{CLEVRER} dataset. The model trained with two frames is consistent, while the model w/o reference changes the type of green object and does not model motion correctly. The green object hits the blue cube and then comes back to hit it again (last frames of the picture). 
    }
    \label{fig:no_reference}
\end{figure*}

\begin{table}[t]
    \centering
    \begin{tabular*}{\linewidth}{l@{\extracolsep{\fill}}cc@{}}
    \toprule
        Method & FVD$\downarrow$ & PSNR$\uparrow$\\
        \hline
        w/ reference & 94.38 & 30.53\\
        w/o reference & 217.13 & 26.95\\
    \bottomrule
    \end{tabular*}
    \caption{Ablations on the use of the reference frame. We generate 14 frames given 2 initial ones and the metrics are calculated on 256 test videos with 1 sample per video and 10 integration steps per frame. All models are trained for 80K iterations.}
    \label{tab:no_reference}
\end{table}

\subsection{Ablations}
\label{sec:abl}

In this section, we ablate several design choices in order to 
illustrate their impact on the performance of \methodName.

First, we ablate the importance of using the reference frame in the condition. In \cite{watson2022novel}, where the stochastic conditioning was first introduced, only one view from the memory was used at each denoising step for generating a novel view. However, conditioning on one frame from the past does not work for video prediction, since one frame does not contain any information about pre-existing motion. We train a model, where we remove the reference frame from the condition and compare its performance to the full model. For this ablation we test \methodName on the CLEVRER~\cite{Yi2020CLEVRERCE} dataset. We found that without the reference frame in the condition the model is confused about the direction of the motion, which results in jumping objects (see Figure~\ref{fig:no_reference}). For the quantitative results, check Table~\ref{tab:no_reference}.


Given a model trained so that the context frames are sampled from the whole past of a sequence, at inference time we ablate the size of the past window used for the context frames to better understand the impact of the history on the video generation performance. In this ablation, we uniformly sample the context frames from $\{\tau~-~1~-~k, \dots, \tau~-~2\}$ for $k = 2, 4, 6, 8$, and show which past frames better support \methodName's predictions. For this experiment we use our trained model on the BAIR~\cite{Ebert2017SelfSupervisedVP} \sepehr{and KTH~\cite{schuldt2004recognizing}} datasets. Since there are occlusions in BAIR, we suspect that having more context can help to predict the future frames more accurately. \sepehr{Having more context frames also helps to predict a smoother motion for humans in KTH. Table~\ref{tab:horizon_size} shows that there is a trade-off in context size and although} having more context can be useful, on simple datasets having only \sepehr{a few} frames is better to solve the prediction task.

\begin{table}[t]
    \centering
    \begin{tabular*}{\linewidth}{l@{\extracolsep{\fill}}cc}
    \toprule
        Context & BAIR / PSNR$\uparrow$ & \aram{KTH / PSNR$\uparrow$}\\
        \hline
        2 frames & 25.64 & 28.53 \\
        4 frames & 25.94 & 29.07 \\
        6 frames & 26.00 & 30.17 \\
        8 frames & 25.28 & 29.40 \\
    \bottomrule
    \end{tabular*}
    \caption{Ablations on the context size. Using a pretrained model on BAIR~\cite{Ebert2017SelfSupervisedVP} \aram{and KTH~\cite{schuldt2004recognizing}} we observe a trade-off wrt the number of conditioning frames. We believe that datasets with more challenging scenes and dynamics may require more context frames.}
    \label{tab:horizon_size}
\end{table}

Finally we show in Figure~\ref{fig:s_tradeoff} that \warm can be used to generate samples faster (with fewer integration steps) but with a cost on quality. Interestingly we observed that a small speed up factor actually helps the sampling and despite having fewer integration steps leads to better performance. We suspect that this effect is similar to the truncation trick~\cite{Marchesi2017MegapixelSI,Brock2019LargeSG} in GANs. Notice however, that compared to other diffusion-based video generation approaches, \methodName conditions only on 2 past frames for a single neural function evaluation (NFE). Hence, a single NFE is generally less expensive. For instance, it takes $9.97$ seconds for \methodName to generate 16 frames video, while RaMViD~\cite{hoppe2022diffusion} requires $40.47$ seconds with a vanilla scheduler on a single Nvidia GeForce RTX 3090 GPU (on BAIR with 64$\times$64 resolution). For more results, see the supplementary material.

\section{Conclusion}
\label{sec:conclusion}

In this paper we have introduced \methodName, a novel training procedure and a model for video prediction that are based on the recently proposed Flow Matching for image synthesis. We have adapted the latter to videos and incorporated conditioning on an arbitrarily large past frames window through randomly sampling a new context frame at each integration step of the learned flow. Moreover, working in the latent space of a pretrained VQGAN enabled the generation of high-resolution videos. All these have resulted in a simple and effective training procedure, which we hope future works on video synthesis will largely benefit from. We have tested \methodName on several video datasets and found that it is not only able to \sepehr{predict} high-quality videos, but is also flexible enough to be trained to perform other tasks, such as visual planning \sepehr{and video generation.}

\noindent\textbf{Acknowledgements.}
This work was supported by grant 188690 of the Swiss National Science Foundation.

\appendix

\section{Appendix}
In the main paper we have \aram{introduced} \methodName - a new model and \aram{an efficient} training procedure to perform video prediction based on Flow Matching and randomized past frame conditioning.
This supplementary material provides details that could not be included in the main paper due to space limitations.
In section~\ref{sec:arch} we describe in details the architecture of our model and how we trained it on different datasets.
In section~\ref{sec:curve} we show the training curve of the model and in section~\ref{sec:train} we conduct an analysis on the training time and memory consumption and compare with that of other methods.
In section~\ref{sec:samples} we provide more samples generated with our model.

\section{Architecture and Training Details}
\label{sec:arch}

\noindent\textbf{Autoencoder.} In this section we provide the configurations of the VQGAN~\cite{esser2021taming} for all the datasets used in the main paper (see Table~\ref{tab:vqgan}). All models were trained using the code from the official \texttt{taming transformers} repository.\footnote{\url{https://github.com/CompVis/taming-transformers}} 

\noindent\textbf{Vector Field Regressor.} In this section we provide implementation details of the network that regresses the conditional time-dependent vector field $v_t(x \,|\, x^{\tau-1}, x^c, \tau - c)$. As mentioned in the main paper, the network is implemented as a U-ViT~\cite{bao2022all}. The detailed architecture is provided in Figure~\ref{fig:arch} and is shared across all datasets. First, the inputs $x, x^{\tau - 1}$ and $x^c$ are channel-wise concatenated and linearly projected to the inner dimension of the ViT blocks. Besides in and out projection layers, the network consists of 13 standard ViT blocks with 4 long skip connections between the first 4 and the last 4 blocks. At each skip connection the inputs are channel-wise concatenated and projected to the inner dimension of the ViT blocks. All ViT blocks apply layer normalization~\cite{Ba2016LayerN} before the multihead self-attention~\cite{NIPS2017_3f5ee243} (MHSA) layer and the MLP. The inner dimension of all ViT blocks is 768 and 8 heads are used in all MHSA layers.
\begin{table*}[t]
    \centering
    \begin{tabular}{@{}r|cccc}
    \toprule
         & BAIR64$\times$64~\cite{Ebert2017SelfSupervisedVP} & BAIR256$\times$256~\cite{Ebert2017SelfSupervisedVP} & KTH~\cite{schuldt2004recognizing} & CLEVRER~\cite{Yi2020CLEVRERCE} \\
    \hline
        embed\textunderscore dim & 4 & 8 & 4 & 4 \\
        n\textunderscore embed & 16384 & 16384 & 16384 & 8192 \\
        double\textunderscore z & False & False & False & False \\ 
        z\textunderscore channels & 4 & 8 & 4 & 4 \\
        resolution & 64 & 256 & 64 & 128 \\
        in\textunderscore channels & 3 & 3 & 3 & 3 \\
        out\textunderscore ch & 3 & 3 & 3 & 3 \\
        ch & 128 & 128 & 128 & 128 \\
        ch\textunderscore mult & [1,2,2,4] & [1,1,2,2,4] & [1,2,2,4] & [1,2,2,4] \\
        num\textunderscore res\textunderscore blocks & 2 & 2 & 2 & 2\\
        attn\textunderscore resolutions & [16] & [16] & [16] & [16] \\
        dropout & 0.0 & 0.0 & 0.0 & 0.0 \\
    \hline
        disc\textunderscore conditional & False & False & False & - \\
        disc\textunderscore in\textunderscore channels & 3 & 3 & 3 & - \\
        disc\textunderscore start & 20k & 20k & 20k & - \\
        disc\textunderscore weight & 0.8 & 0.8 & 0.8 & - \\
        codebook\textunderscore weight & 1.0 & 1.0 & 1.0 & - \\
    \bottomrule
    \end{tabular}
    \caption{Configurations of VQGAN~\cite{esser2021taming} for different datasets. \aram{Notice that on the CLEVRER~\cite{Yi2020CLEVRERCE} dataset we did not utilize an adversarial training.}}
    \label{tab:vqgan}
\end{table*}

\begin{figure}[t]
    \centering
    \includegraphics[width=\linewidth]{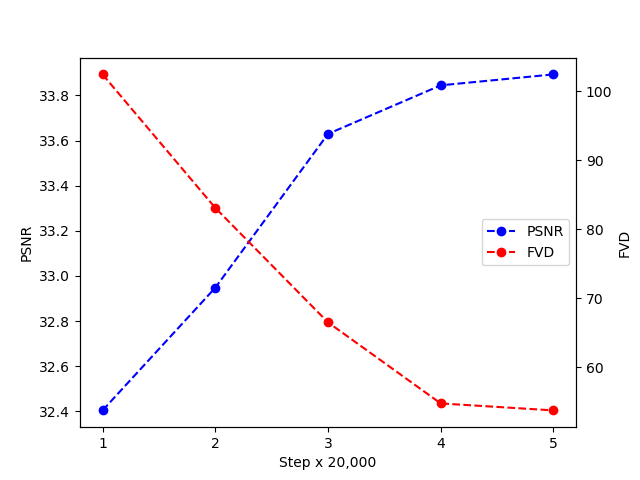}
    \caption{Training curve of \methodName on CLEVRER~\cite{Yi2020CLEVRERCE}.}
    \label{fig:curve}
\end{figure}

\begin{figure*}
    \centering
    \includegraphics[width=\linewidth, trim=2cm 6cm 2cm 10cm, clip]{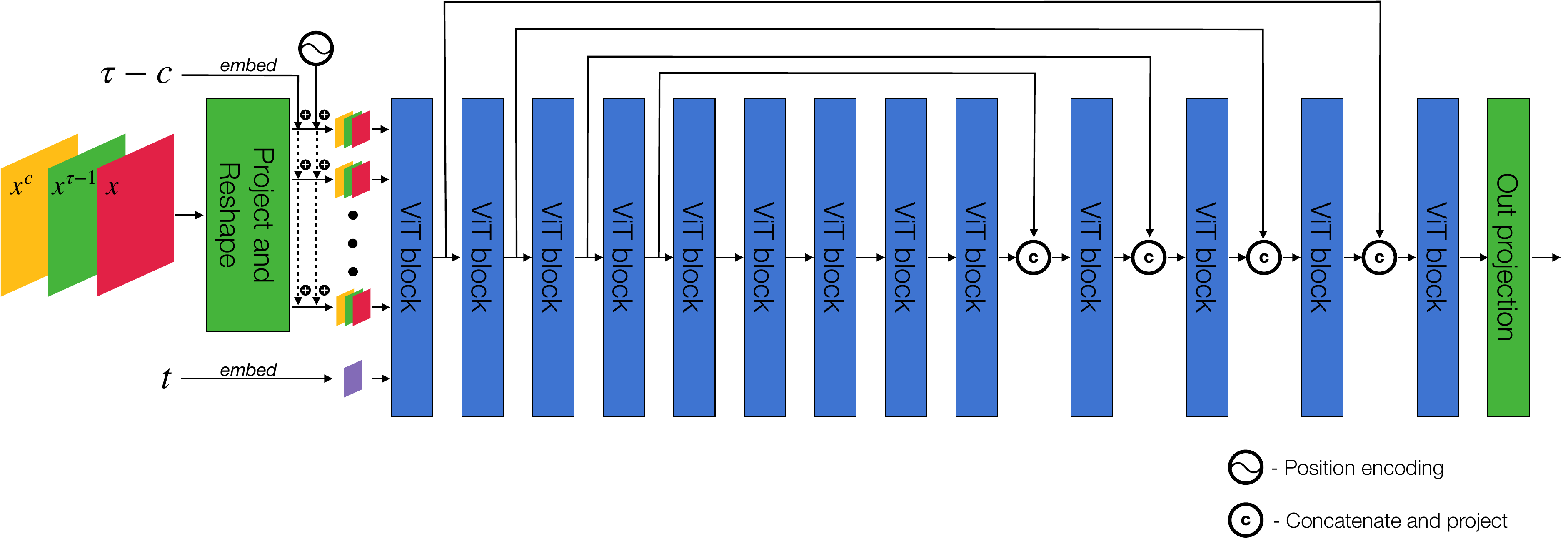}
    \caption{Architecture of the vector field regressor of \methodName. ``ViT block'' stands for a standard self-attention block used in ViT~\cite{Dosovitskiy2021AnII}, that is an MHSA layer, followed by a 2-layer wide MLP, with a layer normalization before each block and a skip connection after each block. ``Out projection'' involves a linear layer, followed by a GELU~\cite{Hendrycks2016GaussianEL} activation, layer normalization and a 3$\times$3 convolutional layer.}
    \label{fig:arch}
\end{figure*}

All models are trained for 300K iterations with the AdamW~\cite{Loshchilov2019DecoupledWD} optimizer with the base learning rate equal to $10^{-4}$ and weight decay $5\cdot 10^{-6}$. A learning rate linear warmup for 5K iterations is used along with a square root decay schedule.
For the CLEVRER~\cite{Yi2020CLEVRERCE} dataset, random color jittering is additionally used to prevent overfitting. We observed that without it, the objects may change colors in the generated sequences (see Figure~\ref{fig:color}).
In all experiments we used $\sigma_{\text{min}} = 10^{-7}$.

Additionally, we would like to highlight once again that the excellent tradeoff of \methodName demonstrated in Figure~1 of the main paper is the motivation to use flow matching instead of diffusion. Flow matching exhibits faster convergence compared to diffusion models. Moreover, on BAIR we observed DDPM fail to converge
(see Figure~\ref{fig:flow_vs_diff}). Besides this, the same theoretical arguments used by the authors of flow matching in the case of images can be extended to the case of videos.

\begin{figure}[h]
    \centering
    \begin{tabular}{cc}
        \animategraphics[width=4cm, loop, autoplay]{7}{Figures/iccv_rebuttal/fm/image_}{0}{15} &
        \animategraphics[width=4cm, loop, autoplay]{7}{Figures/iccv_rebuttal/ddpm/image_}{0}{15} \\
        FM & DDPM \\
    \end{tabular}
    \caption{Video generation with different generative models. Use Acrobat Reader to play videos.}
    \label{fig:flow_vs_diff}
\end{figure}

\begin{figure*}[t]
    \centering
    \begin{tabular}{@{}c@{\hspace{0.5mm}}c@{\hspace{0.5mm}}c@{\hspace{0.5mm}}c@{\hspace{0.5mm}}c@{\hspace{0.5mm}}c@{\hspace{0.5mm}}c@{\hspace{0.5mm}}c@{\hspace{0.5mm}}c@{\hspace{0.5mm}}c@{}}
    \multicolumn{10}{c}{\small time $\rightarrow$}\\
    \animategraphics[width=1.7cm]{7}{Figures/cl_na_gen/000}{0}{18} &
    \includegraphics[width=1.7cm]{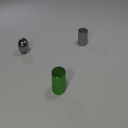} &
    \includegraphics[width=1.7cm]{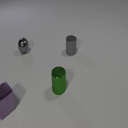} &
    \includegraphics[width=1.7cm]{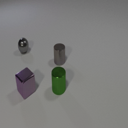} &
    \includegraphics[width=1.7cm]{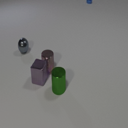} &
    \includegraphics[width=1.7cm]{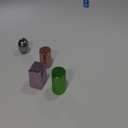} &
    \includegraphics[width=1.7cm]{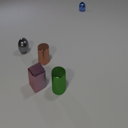} &
    \includegraphics[width=1.7cm]{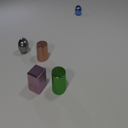} &
    \includegraphics[width=1.7cm]{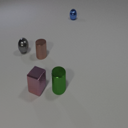} &
    \includegraphics[width=1.7cm]{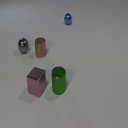} \\
    \end{tabular}
    \caption{A sequence generated with \methodName trained on the \emph{CLEVRER} dataset without data augmentation. Notice how the color of the grey cylinder changes after its interaction with the cube. In order to prevent such behaviour, both the autoencoder and \methodName are trained with random color jittering as data augmentation. \aram{The first frame can be played as a video in Acrobat Reader.}}
    \label{fig:color}
\end{figure*}

\section{Training Curve}
\label{sec:curve}
In Figure~\ref{fig:curve} we show the FVD~\cite{Unterthiner2018TowardsAG} and PSNR of \methodName trained on CLEVRER~\cite{Yi2020CLEVRERCE} against the iteration time. As we can see, the training is stable and more iterations lead to better results.

\section{Training Time and Memory Consumption}
\label{sec:train}
In Table~\ref{tab:time}, we compare the total training time and GPU (or TPU) memory requirements of different models trained on BAIR64$\times$64~\cite{Ebert2017SelfSupervisedVP}. As we can see, \methodName is extremely efficient and can achieve a reasonable FVD~\cite{Unterthiner2018TowardsAG} with significantly less compute than the other methods. For example, SAVP~\cite{lee2018stochastic}, which has the same FVD as \methodName, requires 4.6$\times$ more compute (measured by Mem$\times$Time) and all the models that take less compute than \methodName have FVDs more than 250.

\begin{table*}[t]
    \centering
    \begin{tabular}{lrrrr}
    \toprule
        Method & Memory (GB) & Time (Hours) & \textbf{Mem$\times$Time (GB$\times$Hour)} & FVD~\cite{Unterthiner2018TowardsAG} \\
    \hline
RVD~\cite{yang2022diffusion} &\phantom{00}24 & - & - & 1272 \\
MoCoGAN~\cite{Tulyakov2018MoCoGANDM} &\phantom{00}16 &23  & 368 &\phantom{0}503 \\
SVG-FP~\cite{denton2018stochastic} &\phantom{00}12 &24 & 288 &\phantom{0}315 \\
CDNA~\cite{finn2016unsupervised} &\phantom{00}10 &20 & 200 &\phantom{0}297 \\
SV2P~\cite{Babaeizadeh2018StochasticVV} &\phantom{00}16 &48 & 768 & \phantom{0}263 \\
SRVP~\cite{Franceschi2020StochasticLR} &\phantom{00}36 &\phantom{0}168 & 6048 &\phantom{0}181 \\
VideoFlow~\cite{Kumar2020VideoFlowAC} &\phantom{0}128 &336 & 43008 & \phantom{0}131 \\
LVT~\cite{rakhimov2020latent} &\phantom{0}128 &48 & 6144 &\phantom{0}126 \\
SAVP~\cite{lee2018stochastic} &\phantom{00}32 &144 & 4608 & \phantom{0}116 \\
DVD-GAN-FP~\cite{Clark2019AdversarialVG} &2048 &24 & 49152 & \phantom{0}110 \\
Video Transformer(S)~\cite{weissenborn2019scaling} &\phantom{0}256 &33 & 8448 & \phantom{0}106 \\
TriVD-GAN-FP~\cite{Luc2020TransformationbasedAV} &1024 &280 & 286720 & \phantom{0}103 \\
CCVS(Low res)~\cite{le2021ccvs} &\phantom{0}128 &40 & 5120 & \phantom{00}99 \\
MCVD(spatin)~\cite{voleti2022mcvd} &\phantom{00}86 &50 & 4300 &\phantom{00}97 \\
Video Transformer(L)~\cite{weissenborn2019scaling} &\phantom{0}512 &336 & 172032 & \phantom{00}94 \\
FitVid~\cite{babaeizadeh2021fitvid} &1024 &\phantom{0}288 & 294912 & \phantom{00}94 \\
MCVD(concat)~\cite{voleti2022mcvd} &\phantom{00}77 &78 & 6006 & \phantom{00}90 \\
NUWA~\cite{Wu2022NWAVS} &2560 &336 & 860160 & \phantom{00}87 \\
RaMViD~\cite{hoppe2022diffusion} &\phantom{0}320 &~72 & 23040 & \phantom{00}83 \\
\hline
\methodName &\phantom{00}25 &25 & 625 &\phantom{0}106 \\
    \bottomrule
    \end{tabular}
    \caption{Compute comparisons. We report the memory and training times requirements of different models trained on BAIR64$\times$64~\cite{Ebert2017SelfSupervisedVP}. The overall compute (Mem $\times$ Time) shows that \methodName delivers better FVD with much less compute.}
    \label{tab:time}
\end{table*}

\section{Sampling Speed}
\label{sec:speed}

\begin{figure}
    \centering
    \includegraphics[width=\linewidth]{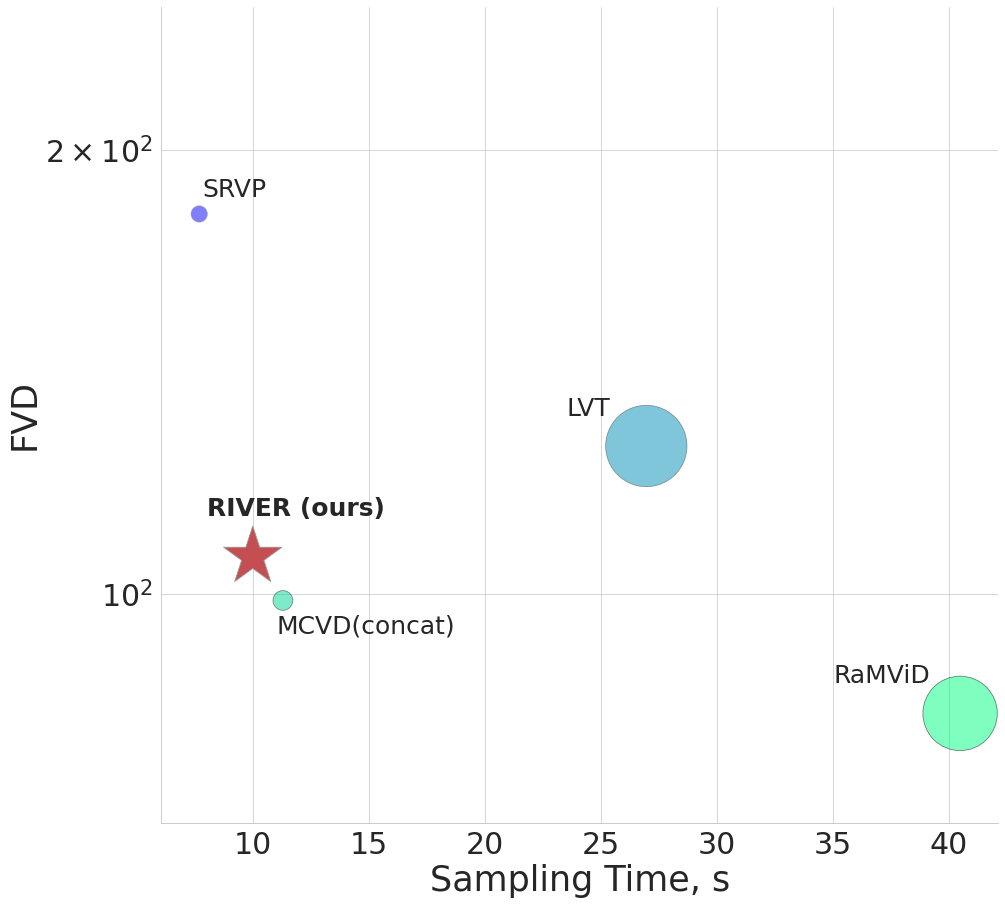}
    \caption{FVD vs. inference speed, the time required to generate a 16 frames long 64$\times$64 resolution video on a single Nvidia GeForce RTX 3090 GPU. The sizes of the markers are proportional to the standard deviation of measured times in 20 independent experiments.}
    \label{fig:tradeoff_sampling}
\end{figure}

In this section we provide more comparisons in terms of the sampling speed with different models. We test the models on the BAIR $64\times 64$ dataset, generating 16 frames and measuring the time the generation required. For evaluation we compare to some diffusion-based models with available code (RaMViD~\cite{hoppe2022diffusion}, MCVD~\cite{voleti2022mcvd}). In addition, we pick one RNN-based model (SRVP~\cite{Franceschi2020StochasticLR}) and one Transformer-based (LVT~\cite{rakhimov2020latent}), to cover different model architectures. The results are reported in Figure~\ref{fig:tradeoff_sampling}. Due to the sparse past frame conditioning, \methodName is able to generate videos with reasonable sampling time. However, if the focus is on the inference speed, one might opt for RNN-based models.

\section{Qualitative Results}
\label{sec:samples}

Here we provide more visual examples of the sequences generated with \methodName. See Figures~\ref{fig:bair} and \ref{fig:bair_fail} for results on the BAIR~\cite{Ebert2017SelfSupervisedVP} dataset, Figures~\ref{fig:kth} and \ref{fig:kth_fail} for results on the KTH~\cite{schuldt2004recognizing} dataset and Figures~\ref{fig:clevrer} and \ref{fig:planning} for video prediction and planning on the CLEVRER~\cite{Yi2020CLEVRERCE} dataset respectively. Besides this, we highlight the stochastic nature of the generation process with \methodName in Figure~\ref{fig:stoch} \aram{and the impact of extreme ($s > 0.5$) \warm strength in Figure~\ref{fig:ws}. 
For more qualitative results and visual comparisons with the prior work, please, visit our website \url{https://araachie.github.io/river}.}

\begin{figure*}[t]
    \centering
    \begin{tabular}{@{}r@{}c@{\hspace{1mm}}c@{\hspace{0.5mm}}c@{\hspace{0.5mm}}c@{\hspace{0.5mm}}c@{\hspace{0.5mm}}c@{\hspace{0.5mm}}c@{\hspace{0.5mm}}c@{\hspace{0.5mm}}c@{}}
    & \makebox[1.7cm][c]{\small last context frame} & \multicolumn{8}{c}{\small time $\rightarrow$}\\
    \rotatebox{90}{\makebox[1.7cm][c]{\small GT}} &
    \multicolumn{1}{c@{\hspace{0.5mm}}?@{\hspace{0.5mm}}}{\includegraphics[width=1.7cm]{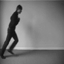}} & 
    \animategraphics[width=1.7cm]{7}{Figures/kth_g6_gt/000}{7}{28} & 
    \includegraphics[width=1.7cm]{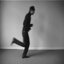} & 
    \includegraphics[width=1.7cm]{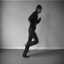} & 
    \includegraphics[width=1.7cm]{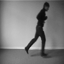} & 
    \includegraphics[width=1.7cm]{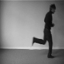} &
    \includegraphics[width=1.7cm]{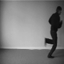} & 
    \includegraphics[width=1.7cm]{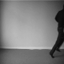} & 
    \includegraphics[width=1.7cm]{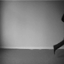} \\
    & \multicolumn{1}{r@{\hspace{0.5mm}}?@{\hspace{0.5mm}}}{\rotatebox{90}{\makebox[1.7cm][c]{predicted}}} &
    \animategraphics[width=1.7cm]{7}{Figures/kth_g6_gen/000}{7}{28} &
    \includegraphics[width=1.7cm]{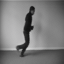} &
    \includegraphics[width=1.7cm]{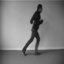} & 
    \includegraphics[width=1.7cm]{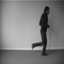} & 
    \includegraphics[width=1.7cm]{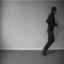} &
    \includegraphics[width=1.7cm]{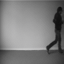} & 
    \includegraphics[width=1.7cm]{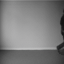} & 
    \includegraphics[width=1.7cm]{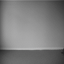} \\
    \rotatebox{90}{\makebox[1.7cm][c]{\small GT}} &
    \multicolumn{1}{c@{\hspace{0.5mm}}?@{\hspace{0.5mm}}}{\includegraphics[width=1.7cm]{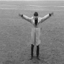}} & 
    \animategraphics[width=1.7cm]{7}{Figures/kth_g5_gt/000}{7}{28} & 
    \includegraphics[width=1.7cm]{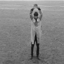} & 
    \includegraphics[width=1.7cm]{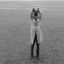} & 
    \includegraphics[width=1.7cm]{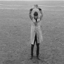} & 
    \includegraphics[width=1.7cm]{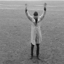} &
    \includegraphics[width=1.7cm]{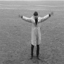} & 
    \includegraphics[width=1.7cm]{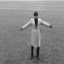} & 
    \includegraphics[width=1.7cm]{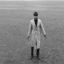} \\
    & \multicolumn{1}{r@{\hspace{0.5mm}}?@{\hspace{0.5mm}}}{\rotatebox{90}{\makebox[1.7cm][c]{predicted}}} &
    \animategraphics[width=1.7cm]{7}{Figures/kth_g5_gen/000}{7}{28} &
    \includegraphics[width=1.7cm]{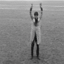} &
    \includegraphics[width=1.7cm]{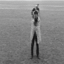} & 
    \includegraphics[width=1.7cm]{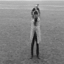} & 
    \includegraphics[width=1.7cm]{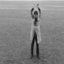} &
    \includegraphics[width=1.7cm]{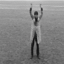} & 
    \includegraphics[width=1.7cm]{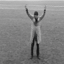} & 
    \includegraphics[width=1.7cm]{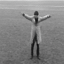} \\
    \rotatebox{90}{\makebox[1.3cm][c]{\small GT}} & \multicolumn{1}{c@{\hspace{0.5mm}}?@{\hspace{0.5mm}}}{\includegraphics[width=1.7cm]{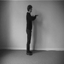}} & 
    \animategraphics[width=1.7cm]{7}{Figures/kth_g3_gt/000}{7}{28} & 
    \includegraphics[width=1.7cm]{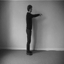} & 
    \includegraphics[width=1.7cm]{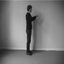} & 
    \includegraphics[width=1.7cm]{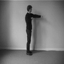} & 
    \includegraphics[width=1.7cm]{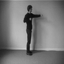} &
    \includegraphics[width=1.7cm]{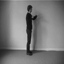} & 
    \includegraphics[width=1.7cm]{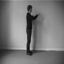} & 
    \includegraphics[width=1.7cm]{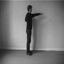} \\
    & \multicolumn{1}{r@{\hspace{0.5mm}}?@{\hspace{0.5mm}}}{\rotatebox{90}{\makebox[1.7cm][c]{predicted}}} &
    \animategraphics[width=1.7cm]{7}{Figures/kth_g3_gen/000}{7}{28} & 
    \includegraphics[width=1.7cm]{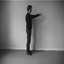} & 
    \includegraphics[width=1.7cm]{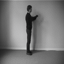} & 
    \includegraphics[width=1.7cm]{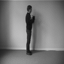} & 
    \includegraphics[width=1.7cm]{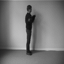} &
    \includegraphics[width=1.7cm]{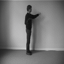} & 
    \includegraphics[width=1.7cm]{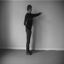} & 
    \includegraphics[width=1.7cm]{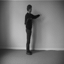} \\
    \rotatebox{90}{\makebox[1.3cm][c]{\small GT}} & \multicolumn{1}{c@{\hspace{0.5mm}}?@{\hspace{0.5mm}}}{\includegraphics[width=1.7cm]{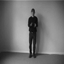}} & 
    \animategraphics[width=1.7cm]{7}{Figures/kth_g4_gt/000}{7}{28} & 
    \includegraphics[width=1.7cm]{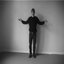} & 
    \includegraphics[width=1.7cm]{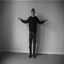} & 
    \includegraphics[width=1.7cm]{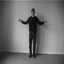} & 
    \includegraphics[width=1.7cm]{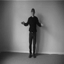} &
    \includegraphics[width=1.7cm]{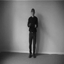} & 
    \includegraphics[width=1.7cm]{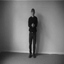} & 
    \includegraphics[width=1.7cm]{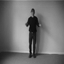} \\
    & \multicolumn{1}{r@{\hspace{0.5mm}}?@{\hspace{0.5mm}}}{\rotatebox{90}{\makebox[1.7cm][c]{predicted}}} &
    \animategraphics[width=1.7cm]{7}{Figures/kth_g4_gen/000}{7}{28} & 
    \includegraphics[width=1.7cm]{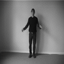} & 
    \includegraphics[width=1.7cm]{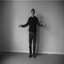} & 
    \includegraphics[width=1.7cm]{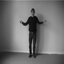} & 
    \includegraphics[width=1.7cm]{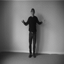} &
    \includegraphics[width=1.7cm]{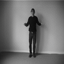} & 
    \includegraphics[width=1.7cm]{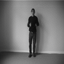} & 
    \includegraphics[width=1.7cm]{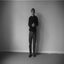} \\
    \end{tabular}
    \caption{Video prediction on the \emph{KTH} dataset. Odd rows show frames of the original video. Even rows show the video generated by \methodName when fed the context frames of the row above (GT). We observe that \methodName is able to generate sequences with diversity and realism. \aram{The images in the first column after the bold vertical line can be played as videos in Acrobat Reader.}}
    \label{fig:kth}
\end{figure*}

\begin{figure*}[t]
    \centering
    \begin{tabular}{@{}r@{}c@{\hspace{1mm}}c@{\hspace{0.5mm}}c@{\hspace{0.5mm}}c@{\hspace{0.5mm}}c@{\hspace{0.5mm}}c@{}}
    & \makebox[2.6cm][c]{last context frame} & \multicolumn{5}{c}{time $\rightarrow$}\\
    \rotatebox{90}{\makebox[2.6cm][c]{GT}} &
    \multicolumn{1}{c@{\hspace{0.5mm}}?@{\hspace{0.5mm}}}{\includegraphics[width=2.6cm]{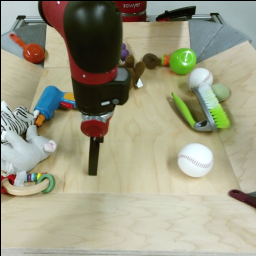}} & 
    \animategraphics[width=2.6cm, loop]{7}{Figures/bair_g1_gt/0000}{1}{5} &
    \includegraphics[width=2.6cm]{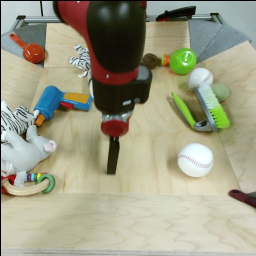} &
    \includegraphics[width=2.6cm]{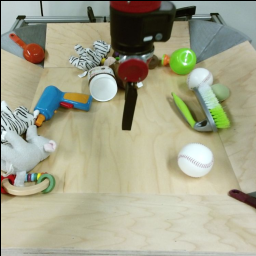} &
    \includegraphics[width=2.6cm]{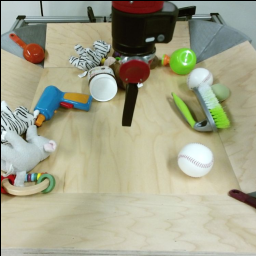} &
    \includegraphics[width=2.6cm]{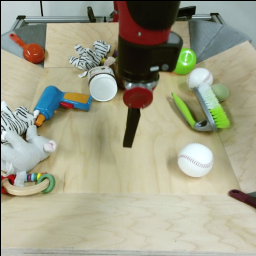} \\
    & \multicolumn{1}{r@{\hspace{0.5mm}}?@{\hspace{0.5mm}}}{\rotatebox{90}{\makebox[2.6cm][c]{predicted}}} & 
    \animategraphics[width=2.6cm, loop]{7}{Figures/bair_g1_gen/0000}{1}{5} &
    \includegraphics[width=2.6cm]{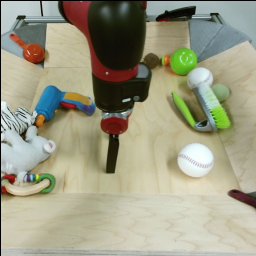} &
    \includegraphics[width=2.6cm]{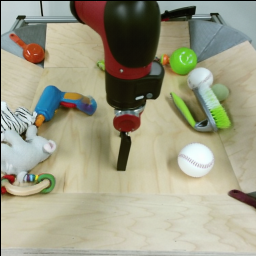} &
    \includegraphics[width=2.6cm]{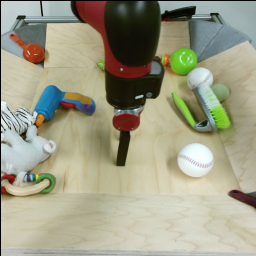} &
    \includegraphics[width=2.6cm]{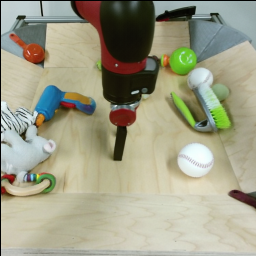}\\
    \rotatebox{90}{\makebox[2.6cm][c]{GT}} &
    \multicolumn{1}{c@{\hspace{0.5mm}}?@{\hspace{0.5mm}}}{\includegraphics[width=2.6cm]{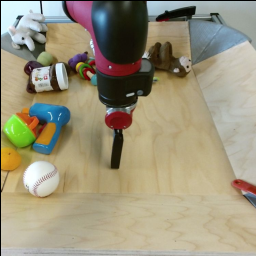}} & 
    \animategraphics[width=2.6cm, loop]{7}{Figures/bair_g2_gt/0000}{1}{5} &
    \includegraphics[width=2.6cm]{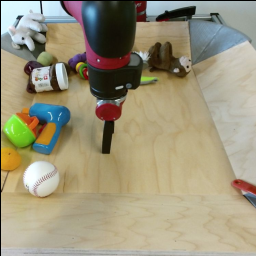} &
    \includegraphics[width=2.6cm]{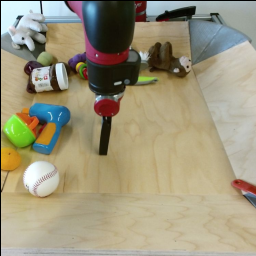} &
    \includegraphics[width=2.6cm]{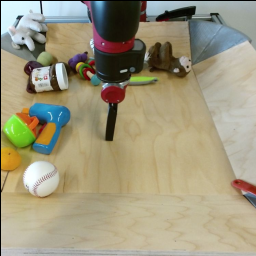} &
    \includegraphics[width=2.6cm]{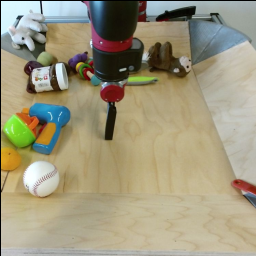} \\
    & \multicolumn{1}{r@{\hspace{0.5mm}}?@{\hspace{0.5mm}}}{\rotatebox{90}{\makebox[2.6cm][c]{predicted}}} & 
    \animategraphics[width=2.6cm, loop]{7}{Figures/bair_g2_gen/0000}{1}{5} &
    \includegraphics[width=2.6cm]{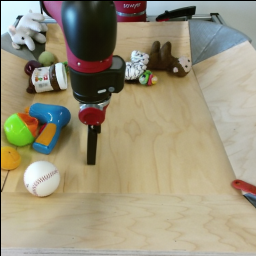} &
    \includegraphics[width=2.6cm]{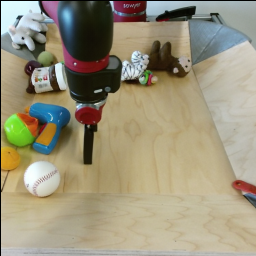} &
    \includegraphics[width=2.6cm]{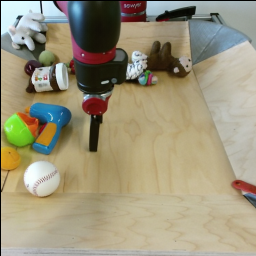} &
    \includegraphics[width=2.6cm]{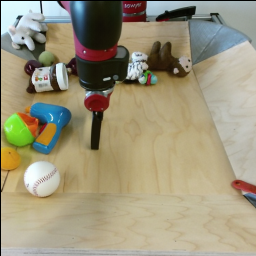}\\
    \rotatebox{90}{\makebox[2.6cm][c]{GT}} &
    \multicolumn{1}{c@{\hspace{0.5mm}}?@{\hspace{0.5mm}}}{\includegraphics[width=2.6cm]{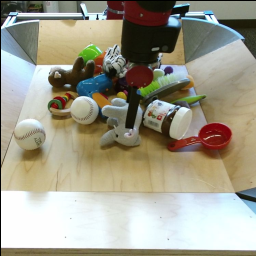}} & 
    \animategraphics[width=2.6cm, loop]{7}{Figures/bair_g3_gt/0000}{1}{5} &
    \includegraphics[width=2.6cm]{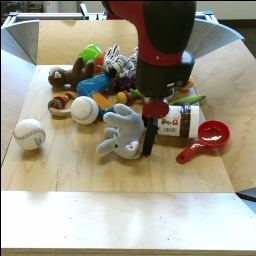} &
    \includegraphics[width=2.6cm]{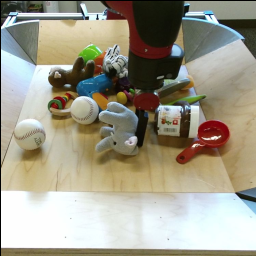} &
    \includegraphics[width=2.6cm]{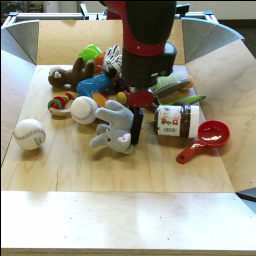} &
    \includegraphics[width=2.6cm]{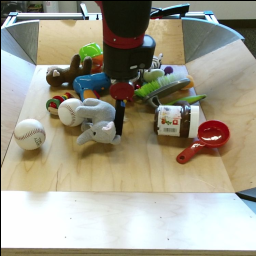} \\
    & \multicolumn{1}{r@{\hspace{0.5mm}}?@{\hspace{0.5mm}}}{\rotatebox{90}{\makebox[2.6cm][c]{predicted}}} & 
    \animategraphics[width=2.6cm, loop]{7}{Figures/bair_g3_gen/0000}{1}{5} &
    \includegraphics[width=2.6cm]{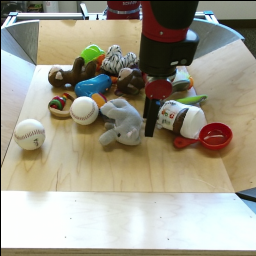} &
    \includegraphics[width=2.6cm]{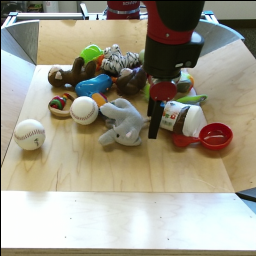} &
    \includegraphics[width=2.6cm]{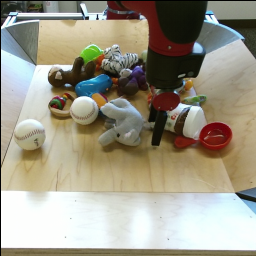} &
    \includegraphics[width=2.6cm]{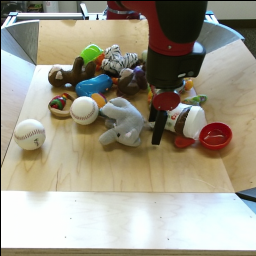}\\
    \rotatebox{90}{\makebox[2.6cm][c]{GT}} &
    \multicolumn{1}{c@{\hspace{0.5mm}}?@{\hspace{0.5mm}}}{\includegraphics[width=2.6cm]{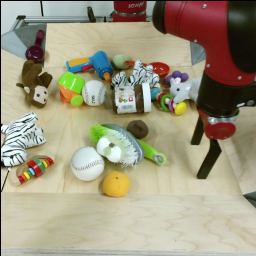}} & 
    \animategraphics[width=2.6cm, loop]{7}{Figures/bair_g4_gt/0000}{1}{5} &
    \includegraphics[width=2.6cm]{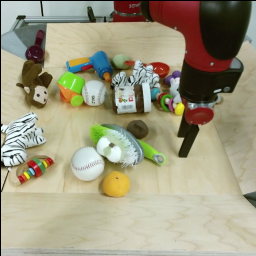} &
    \includegraphics[width=2.6cm]{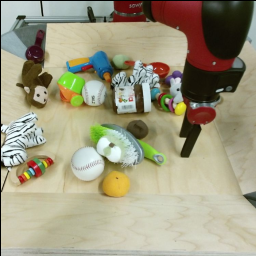} &
    \includegraphics[width=2.6cm]{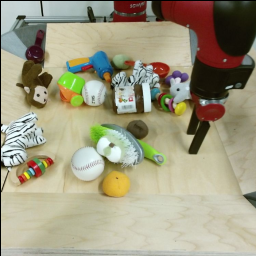} &
    \includegraphics[width=2.6cm]{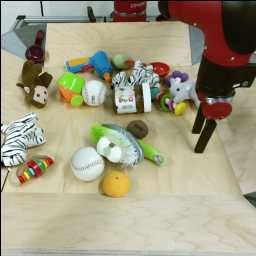} \\
    & \multicolumn{1}{r@{\hspace{0.5mm}}?@{\hspace{0.5mm}}}{\rotatebox{90}{\makebox[2.6cm][c]{predicted}}} & 
    \animategraphics[width=2.6cm, loop]{7}{Figures/bair_g4_gen/0000}{1}{5} &
    \includegraphics[width=2.6cm]{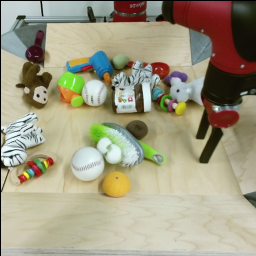} &
    \includegraphics[width=2.6cm]{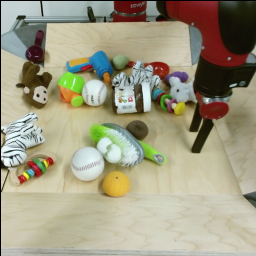} &
    \includegraphics[width=2.6cm]{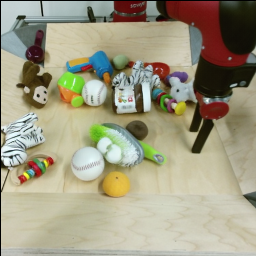} &
    \includegraphics[width=2.6cm]{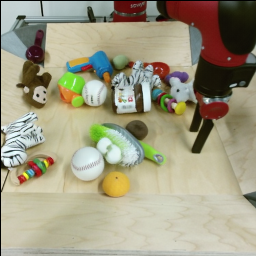}\\
    \end{tabular}
    \caption{Video prediction on the \emph{BAIR} dataset \aram{at $256\times 256$ resolution}. The model predicts the future frames conditioned on a single initial frame. \aram{The frames in the first column after the bold vertical line can be played as videos in Acrobat Reader.}}
    \label{fig:bair}
\end{figure*}

\begin{figure*}[t]
    \centering
    \begin{tabular}{@{}r@{}c@{\hspace{1mm}}c@{\hspace{0.5mm}}c@{\hspace{0.5mm}}c@{\hspace{0.5mm}}c@{\hspace{0.5mm}}c@{\hspace{0.5mm}}c@{\hspace{0.5mm}}c@{\hspace{0.5mm}}c@{}}
    & \makebox[1.7cm][c]{\small last context frame} & \multicolumn{8}{c}{\small time $\rightarrow$}\\
    \rotatebox{90}{\makebox[1.3cm][c]{\small GT}} & \multicolumn{1}{c@{\hspace{0.5mm}}?@{\hspace{0.5mm}}}{\includegraphics[width=1.7cm]{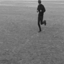}} & 
    \animategraphics[width=1.7cm]{7}{Figures/kth_f1_gt/000}{6}{27} & 
    \includegraphics[width=1.7cm]{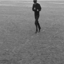} & 
    \includegraphics[width=1.7cm]{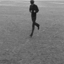} & 
    \includegraphics[width=1.7cm]{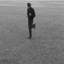} & 
    \includegraphics[width=1.7cm]{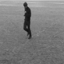} &
    \includegraphics[width=1.7cm]{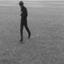} & 
    \includegraphics[width=1.7cm]{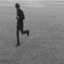} & 
    \includegraphics[width=1.7cm]{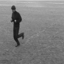} \\
    & \multicolumn{1}{r@{\hspace{0.5mm}}?@{\hspace{0.5mm}}}{\rotatebox{90}{\makebox[1.7cm][c]{predicted}}} &
    \animategraphics[width=1.7cm]{7}{Figures/kth_f1_gen/000}{6}{27} & 
    \includegraphics[width=1.7cm]{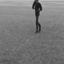} & 
    \includegraphics[width=1.7cm]{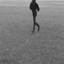} & 
    \includegraphics[width=1.7cm]{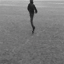} & 
    \includegraphics[width=1.7cm]{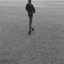} &
    \includegraphics[width=1.7cm]{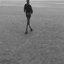} & 
    \includegraphics[width=1.7cm]{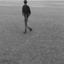} & 
    \includegraphics[width=1.7cm]{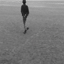} \\
    \rotatebox{90}{\makebox[1.3cm][c]{\small GT}} & \multicolumn{1}{c@{\hspace{0.5mm}}?@{\hspace{0.5mm}}}{\includegraphics[width=1.7cm]{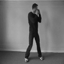}} & 
    \animategraphics[width=1.7cm]{7}{Figures/kth_f2_gt/000}{6}{27} & 
    \includegraphics[width=1.7cm]{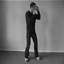} & 
    \includegraphics[width=1.7cm]{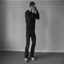} & 
    \includegraphics[width=1.7cm]{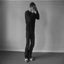} & 
    \includegraphics[width=1.7cm]{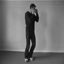} &
    \includegraphics[width=1.7cm]{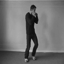} & 
    \includegraphics[width=1.7cm]{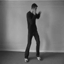} & 
    \includegraphics[width=1.7cm]{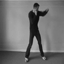} \\
    & \multicolumn{1}{r@{\hspace{0.5mm}}?@{\hspace{0.5mm}}}{\rotatebox{90}{\makebox[1.7cm][c]{predicted}}} &
    \animategraphics[width=1.7cm]{7}{Figures/kth_f2_gen/000}{6}{27} & 
    \includegraphics[width=1.7cm]{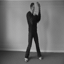} & 
    \includegraphics[width=1.7cm]{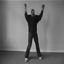} & 
    \includegraphics[width=1.7cm]{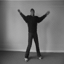} & 
    \includegraphics[width=1.7cm]{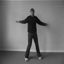} &
    \includegraphics[width=1.7cm]{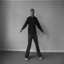} & 
    \includegraphics[width=1.7cm]{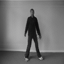} & 
    \includegraphics[width=1.7cm]{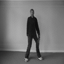} \\
    \end{tabular}
    \caption{Failure cases on the \emph{KTH} dataset. A common failure mode is when a certain action gets confused with another one, which results in a motion that morphs into a different one. In all examples the model is asked to predict 25 future frames given the first 5. \aram{The images in the first column after the bold vertical line can be played as videos in Acrobat Reader.}}
    \label{fig:kth_fail}
\end{figure*}

\begin{figure*}[t]
    \centering
    \begin{tabular}{@{}r@{}c@{\hspace{1mm}}c@{\hspace{0.5mm}}c@{\hspace{0.5mm}}c@{\hspace{0.5mm}}c@{\hspace{0.5mm}}c@{}}
    & \makebox[2.6cm][c]{last context frame} & \multicolumn{5}{c}{time $\rightarrow$}\\
    \rotatebox{90}{\makebox[2.6cm][c]{GT}} &
    \multicolumn{1}{c@{\hspace{0.5mm}}?@{\hspace{0.5mm}}}{\includegraphics[width=2.6cm]{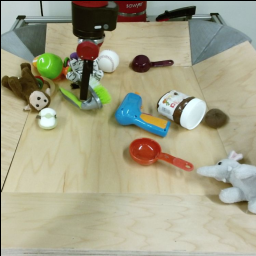}} & 
    \animategraphics[width=2.6cm]{7}{Figures/bair_f1_gt/000}{3}{15} &
    \includegraphics[width=2.6cm]{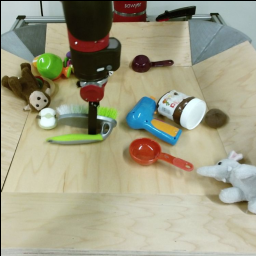} &
    \includegraphics[width=2.6cm]{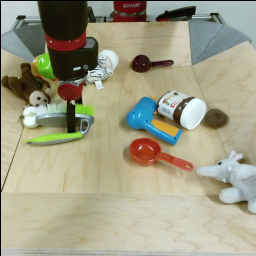} &
    \includegraphics[width=2.6cm]{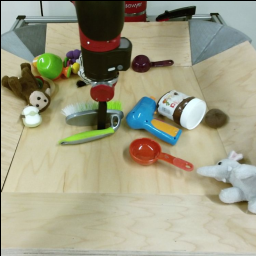} &
    \includegraphics[width=2.6cm]{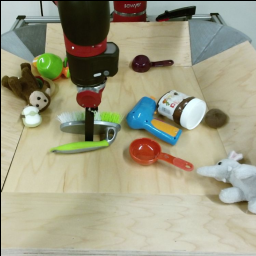} \\
    & \multicolumn{1}{r@{\hspace{0.5mm}}?@{\hspace{0.5mm}}}{\rotatebox{90}{\makebox[2.6cm][c]{predicted}}} & 
    \animategraphics[width=2.6cm]{7}{Figures/bair_f1_gen/000}{3}{15} &
    \includegraphics[width=2.6cm]{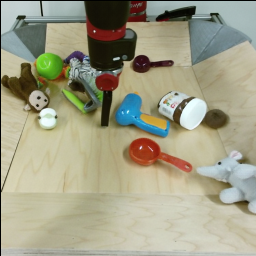} &
    \includegraphics[width=2.6cm]{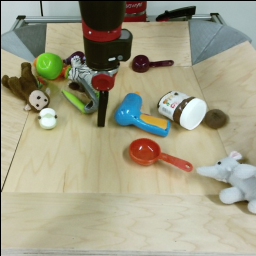} &
    \includegraphics[width=2.6cm]{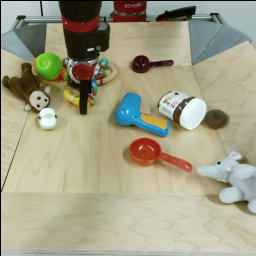} &
    \includegraphics[width=2.6cm]{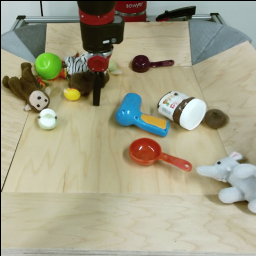}\\
    \end{tabular}
    \caption{Failure case on the \emph{BAIR} dataset. A common failure mode \aram{emerges when generating longer sequences and} is when the interaction causes objects to change their class, shape or even to disappear. \aram{The images in the first column after the bold vertical line can be played as videos in Acrobat Reader.}}
    \label{fig:bair_fail}
\end{figure*}

\begin{figure*}[t]
    \centering
    \begin{tabular}{@{}r@{}c@{\hspace{1mm}}c@{\hspace{0.5mm}}c@{\hspace{0.5mm}}c@{\hspace{0.5mm}}c@{\hspace{0.5mm}}c@{\hspace{0.5mm}}c@{\hspace{0.5mm}}c@{\hspace{0.5mm}}c@{}}
    & \makebox[1.7cm][c]{\small last context frame} & \multicolumn{8}{c}{\small time $\rightarrow$}\\
    \rotatebox{90}{\makebox[1.7cm][c]{\small GT}} &
    \multicolumn{1}{c@{\hspace{0.5mm}}?@{\hspace{0.5mm}}}{\includegraphics[width=1.7cm]{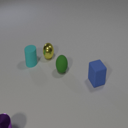}} &
    \animategraphics[width=1.7cm]{7}{Figures/cl_g1_gt/000}{15}{22} & 
    \includegraphics[width=1.7cm]{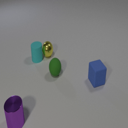} & 
    \includegraphics[width=1.7cm]{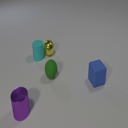} & 
    \includegraphics[width=1.7cm]{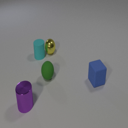} & 
    \includegraphics[width=1.7cm]{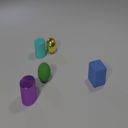} &
    \includegraphics[width=1.7cm]{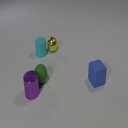} & 
    \includegraphics[width=1.7cm]{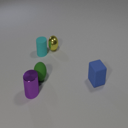} & 
    \includegraphics[width=1.7cm]{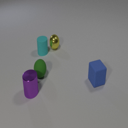} \\
    & \multicolumn{1}{r@{\hspace{0.5mm}}?@{\hspace{0.5mm}}}{\rotatebox{90}{\makebox[1.7cm][c]{predicted}}} &
    \animategraphics[width=1.7cm]{7}{Figures/cl_g1_gen/0000}{2}{9} &
    \includegraphics[width=1.7cm]{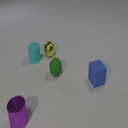} &
    \includegraphics[width=1.7cm]{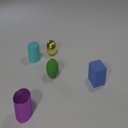} & 
    \includegraphics[width=1.7cm]{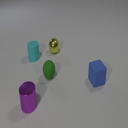} & 
    \includegraphics[width=1.7cm]{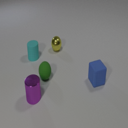} &
    \includegraphics[width=1.7cm]{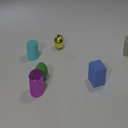} & 
    \includegraphics[width=1.7cm]{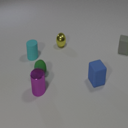} & 
    \includegraphics[width=1.7cm]{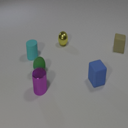} \\
    \rotatebox{90}{\makebox[1.3cm][c]{\small GT}} & \multicolumn{1}{c@{\hspace{0.5mm}}?@{\hspace{0.5mm}}}{\includegraphics[width=1.7cm]{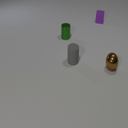}} & 
    \animategraphics[width=1.7cm]{7}{Figures/cl_g2_gt/000}{2}{19} & 
    \includegraphics[width=1.7cm]{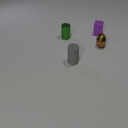} & 
    \includegraphics[width=1.7cm]{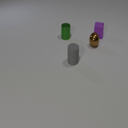} & 
    \includegraphics[width=1.7cm]{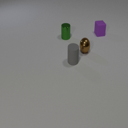} & 
    \includegraphics[width=1.7cm]{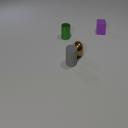} &
    \includegraphics[width=1.7cm]{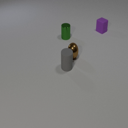} & 
    \includegraphics[width=1.7cm]{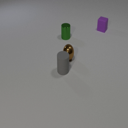} & 
    \includegraphics[width=1.7cm]{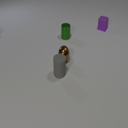} \\
    & \multicolumn{1}{r@{\hspace{0.5mm}}?@{\hspace{0.5mm}}}{\rotatebox{90}{\makebox[1.7cm][c]{predicted}}} &
    \animategraphics[width=1.7cm]{7}{Figures/cl_g2_gen/000}{2}{19} & 
    \includegraphics[width=1.7cm]{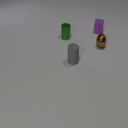} & 
    \includegraphics[width=1.7cm]{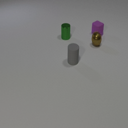} & 
    \includegraphics[width=1.7cm]{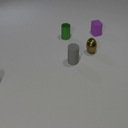} & 
    \includegraphics[width=1.7cm]{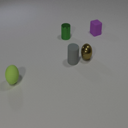} &
    \includegraphics[width=1.7cm]{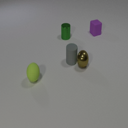} & 
    \includegraphics[width=1.7cm]{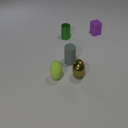} & 
    \includegraphics[width=1.7cm]{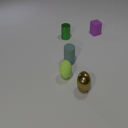} \\
    \rotatebox{90}{\makebox[1.3cm][c]{\small GT}} & \multicolumn{1}{c@{\hspace{0.5mm}}?@{\hspace{0.5mm}}}{\includegraphics[width=1.7cm]{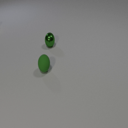}} & 
    \animategraphics[width=1.7cm]{7}{Figures/cl_g3_gt/000}{2}{19} & 
    \includegraphics[width=1.7cm]{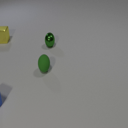} & 
    \includegraphics[width=1.7cm]{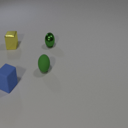} & 
    \includegraphics[width=1.7cm]{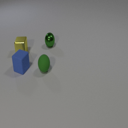} & 
    \includegraphics[width=1.7cm]{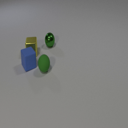} &
    \includegraphics[width=1.7cm]{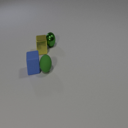} & 
    \includegraphics[width=1.7cm]{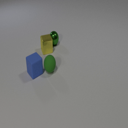} & 
    \includegraphics[width=1.7cm]{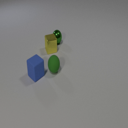} \\
    & \multicolumn{1}{r@{\hspace{0.5mm}}?@{\hspace{0.5mm}}}{\rotatebox{90}{\makebox[1.7cm][c]{predicted}}} &
    \animategraphics[width=1.7cm]{7}{Figures/cl_g3_gen/000}{2}{19} & 
    \includegraphics[width=1.7cm]{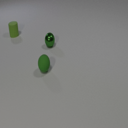} & 
    \includegraphics[width=1.7cm]{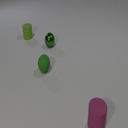} & 
    \includegraphics[width=1.7cm]{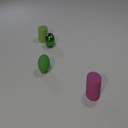} & 
    \includegraphics[width=1.7cm]{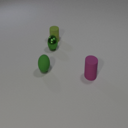} &
    \includegraphics[width=1.7cm]{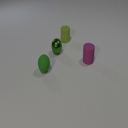} & 
    \includegraphics[width=1.7cm]{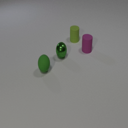} & 
    \includegraphics[width=1.7cm]{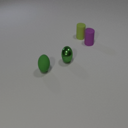} \\
    \rotatebox{90}{\makebox[1.3cm][c]{\small GT}} & \multicolumn{1}{c@{\hspace{0.5mm}}?@{\hspace{0.5mm}}}{\includegraphics[width=1.7cm]{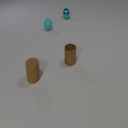}} & 
    \animategraphics[width=1.7cm]{7}{Figures/cl_g4_gt/000}{2}{19} & 
    \includegraphics[width=1.7cm]{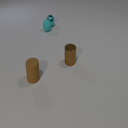} & 
    \includegraphics[width=1.7cm]{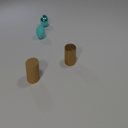} & 
    \includegraphics[width=1.7cm]{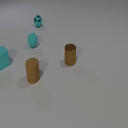} & 
    \includegraphics[width=1.7cm]{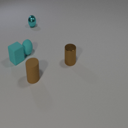} &
    \includegraphics[width=1.7cm]{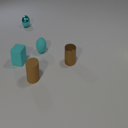} & 
    \includegraphics[width=1.7cm]{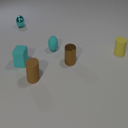} & 
    \includegraphics[width=1.7cm]{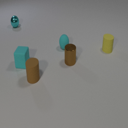} \\
    & \multicolumn{1}{r@{\hspace{0.5mm}}?@{\hspace{0.5mm}}}{\rotatebox{90}{\makebox[1.7cm][c]{predicted}}} &
    \animategraphics[width=1.7cm]{7}{Figures/cl_g4_gen/000}{2}{19} & 
    \includegraphics[width=1.7cm]{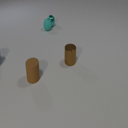} & 
    \includegraphics[width=1.7cm]{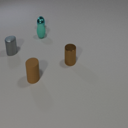} & 
    \includegraphics[width=1.7cm]{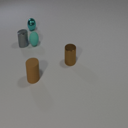} & 
    \includegraphics[width=1.7cm]{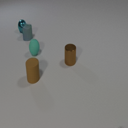} &
    \includegraphics[width=1.7cm]{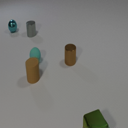} & 
    \includegraphics[width=1.7cm]{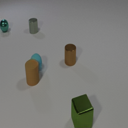} & 
    \includegraphics[width=1.7cm]{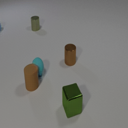} \\
    \end{tabular}
    \caption{Video prediction on the \emph{CLEVRER} dataset. In order to predict the future frames, the model conditions on the first 2 frames. Only the last context frame is shown. The model succeeds to predict the motion that was observed in the context frames. However, it cannot predict new objects as in the ground truth and introduces random new objects due to the stochasticity of the generation process. \aram{The images in the first column after the bold vertical line can be played as videos in Acrobat Reader.}}
    \label{fig:clevrer}
\end{figure*}

\begin{figure*}[t]
    \centering
    \begin{tabular}{@{}c@{\hspace{0.5mm}}c@{\hspace{0.5mm}}c@{\hspace{0.5mm}}c@{\hspace{0.5mm}}c@{\hspace{0.5mm}}c@{\hspace{0.5mm}}c@{\hspace{0.5mm}}c@{\hspace{0.5mm}}c@{\hspace{0.5mm}}c@{}}
    \multicolumn{10}{c}{\small time $\rightarrow$}\\
    \animategraphics[width=1.7cm]{7}{Figures/cl_st1_gen/000}{0}{18} &
    \includegraphics[width=1.7cm]{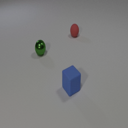} &
    \includegraphics[width=1.7cm]{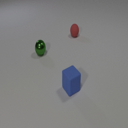} &
    \includegraphics[width=1.7cm]{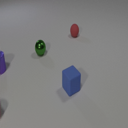} &
    \includegraphics[width=1.7cm]{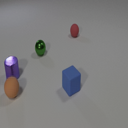} &
    \includegraphics[width=1.7cm]{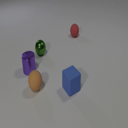} &
    \includegraphics[width=1.7cm]{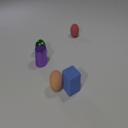} &
    \includegraphics[width=1.7cm]{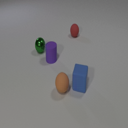} &
    \includegraphics[width=1.7cm]{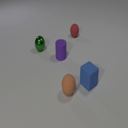} &
    \includegraphics[width=1.7cm]{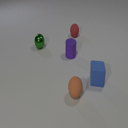} \\
    \animategraphics[width=1.7cm]{7}{Figures/cl_st2_gen/000}{0}{18} &
    \includegraphics[width=1.7cm]{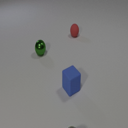} &
    \includegraphics[width=1.7cm]{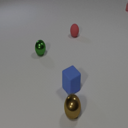} &
    \includegraphics[width=1.7cm]{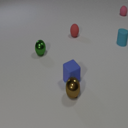} &
    \includegraphics[width=1.7cm]{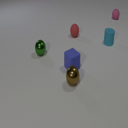} &
    \includegraphics[width=1.7cm]{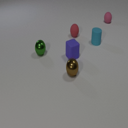} &
    \includegraphics[width=1.7cm]{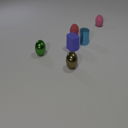} &
    \includegraphics[width=1.7cm]{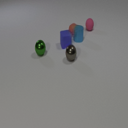} &
    \includegraphics[width=1.7cm]{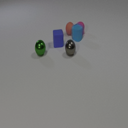} &
    \includegraphics[width=1.7cm]{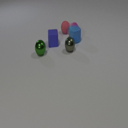} \\
    \end{tabular}
    \caption{Two sequences generated with \methodName trained on the \emph{CLEVRER} dataset. The model was asked to predict 19 frames given 1. Note the very different fates of the blue cube in these two sequences. \aram{The images in the first column can be played as videos in Acrobat Reader.}}
    \label{fig:stoch}
\end{figure*}

\begin{figure*}[t]
    \centering
    \begin{tabular}{@{}c@{\hspace{0.5mm}}c@{\hspace{0.5mm}}c@{\hspace{0.5mm}}c@{\hspace{0.5mm}}c@{\hspace{0.5mm}}c@{}}
    \makebox[2.5cm][c]{source frame} & \multicolumn{3}{c}{time $\rightarrow$} & \makebox[2.5cm][c]{target frame}\\
    \tcbox[hbox, size=fbox, graphics options={width=2.5cm}, colback=green!20,]{\animategraphics[width=2.5cm]{4}{Figures/pl_1_gen/0000}{0}{5}} & 
    \tcbox[hbox, size=fbox, graphics options={width=2.5cm}, colframe=white, colback=white!30,]{\includegraphics[width=2.5cm]{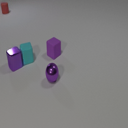}} &
    \tcbox[hbox, size=fbox, graphics options={width=2.5cm}, colframe=white, colback=white!30,]{\includegraphics[width=2.5cm]{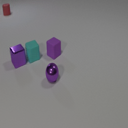}} &
    \tcbox[hbox, size=fbox, graphics options={width=2.5cm}, colframe=white, colback=white!30,]{\includegraphics[width=2.5cm]{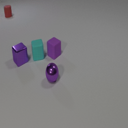}} &
    \tcbox[hbox, size=fbox, graphics options={width=2.5cm}, colframe=white, colback=white!30,]{\includegraphics[width=2.5cm]{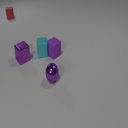}} &
    \tcbox[hbox, size=fbox, graphics options={width=2.5cm}, colback=green!20,]{\includegraphics[width=2.5cm]{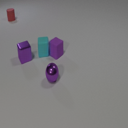}} \\
    \tcbox[hbox, size=fbox, graphics options={width=2.5cm}, colback=green!20,]{\animategraphics[width=2.5cm]{4}{Figures/pl_2_gen/0000}{0}{5}} & 
    \tcbox[hbox, size=fbox, graphics options={width=2.5cm}, colframe=white, colback=white!30,]{\includegraphics[width=2.5cm]{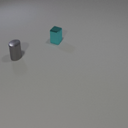}} &
    \tcbox[hbox, size=fbox, graphics options={width=2.5cm}, colframe=white, colback=white!30,]{\includegraphics[width=2.5cm]{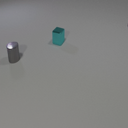}} &
    \tcbox[hbox, size=fbox, graphics options={width=2.5cm}, colframe=white, colback=white!30,]{\includegraphics[width=2.5cm]{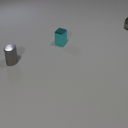}} &
    \tcbox[hbox, size=fbox, graphics options={width=2.5cm}, colframe=white, colback=white!30,]{\includegraphics[width=2.5cm]{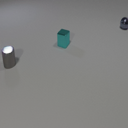}} &
    \tcbox[hbox, size=fbox, graphics options={width=2.5cm}, colback=green!20,]{\includegraphics[width=2.5cm]{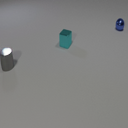}} \\
    \tcbox[hbox, size=fbox, graphics options={width=2.5cm}, colback=green!20,]{\animategraphics[width=2.5cm]{4}{Figures/pl_3_gen/0000}{0}{5}} & 
    \tcbox[hbox, size=fbox, graphics options={width=2.5cm}, colframe=white, colback=white!30,]{\includegraphics[width=2.5cm]{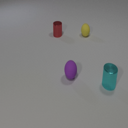}} &
    \tcbox[hbox, size=fbox, graphics options={width=2.5cm}, colframe=white, colback=white!30,]{\includegraphics[width=2.5cm]{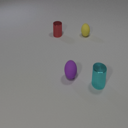}} &
    \tcbox[hbox, size=fbox, graphics options={width=2.5cm}, colframe=white, colback=white!30,]{\includegraphics[width=2.5cm]{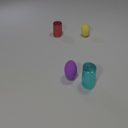}} &
    \tcbox[hbox, size=fbox, graphics options={width=2.5cm}, colframe=white, colback=white!30,]{\includegraphics[width=2.5cm]{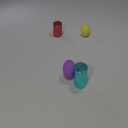}} &
    \tcbox[hbox, size=fbox, graphics options={width=2.5cm}, colback=green!20,]{\includegraphics[width=2.5cm]{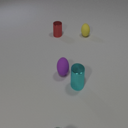}} \\
    \tcbox[hbox, size=fbox, graphics options={width=2.5cm}, colback=green!20,]{\animategraphics[width=2.5cm]{4}{Figures/pl_4_gen/0000}{0}{5}} & 
    \tcbox[hbox, size=fbox, graphics options={width=2.5cm}, colframe=white, colback=white!30,]{\includegraphics[width=2.5cm]{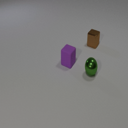}} &
    \tcbox[hbox, size=fbox, graphics options={width=2.5cm}, colframe=white, colback=white!30,]{\includegraphics[width=2.5cm]{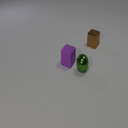}} &
    \tcbox[hbox, size=fbox, graphics options={width=2.5cm}, colframe=white, colback=white!30,]{\includegraphics[width=2.5cm]{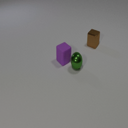}} &
    \tcbox[hbox, size=fbox, graphics options={width=2.5cm}, colframe=white, colback=white!30,]{\includegraphics[width=2.5cm]{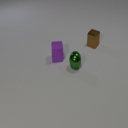}} &
    \tcbox[hbox, size=fbox, graphics options={width=2.5cm}, colback=green!20,]{\includegraphics[width=2.5cm]{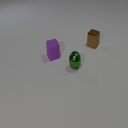}} \\
    \end{tabular}
    \caption{Visual planning with \methodName on the \emph{CLEVRER} dataset. Given the source and the target frames, \methodName generates intermediate frames, so that they form a plausible realistic sequence. \aram{The images in the first column can be played as videos in Acrobat Reader.}}
    \label{fig:planning}
\end{figure*}

\begin{figure*}[t]
    \centering
    \begin{tabular}{@{}r@{\hspace{0.5mm}}c@{\hspace{0.5mm}}c@{\hspace{0.5mm}}c@{\hspace{0.5mm}}c@{\hspace{0.5mm}}c@{\hspace{0.5mm}}c@{\hspace{0.5mm}}c@{\hspace{0.5mm}}c@{\hspace{0.5mm}}c@{\hspace{0.5mm}}c@{}}
    &\multicolumn{10}{c}{\small time $\rightarrow$}\\
    \rotatebox{90}{\makebox[1.5cm][c]{\small $s = 0$}} &
    \animategraphics[width=1.7cm]{7}{Figures/rebuttal/cl_nows_gen/frame_n}{1}{20} &
    \includegraphics[width=1.7cm]{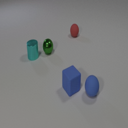} & 
    \includegraphics[width=1.7cm]{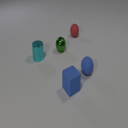} & 
    \includegraphics[width=1.7cm]{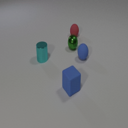} &
    \includegraphics[width=1.7cm]{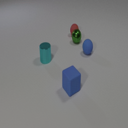} &
    \includegraphics[width=1.7cm]{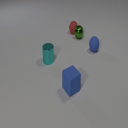} &
    \includegraphics[width=1.7cm]{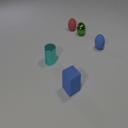} &
    \includegraphics[width=1.7cm]{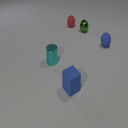} &
    \includegraphics[width=1.7cm]{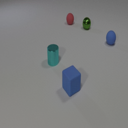} & \includegraphics[width=1.7cm]{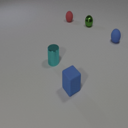} \\
    \rotatebox{90}{\makebox[1.5cm][c]{\small $s = 0.5$}} &
    \animategraphics[width=1.7cm]{7}{Figures/rebuttal/cl_ws05_gen/frame_n}{1}{20} &
    \includegraphics[width=1.7cm]{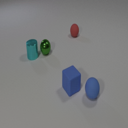} & 
    \includegraphics[width=1.7cm]{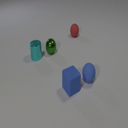} & 
    \includegraphics[width=1.7cm]{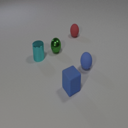} &
    \includegraphics[width=1.7cm]{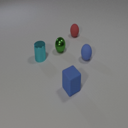} &
    \includegraphics[width=1.7cm]{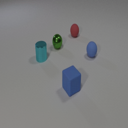} &
    \includegraphics[width=1.7cm]{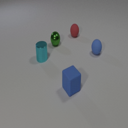} &
    \includegraphics[width=1.7cm]{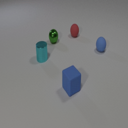} &
    \includegraphics[width=1.7cm]{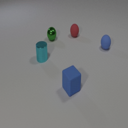} &
    \includegraphics[width=1.7cm]{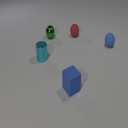} \\
    \end{tabular}
    \caption{\aram{The effect of extreme ($s = 0.5$) \warm strength. The first frame in each row can be played as a video in Acrobat Reader.}}
    \label{fig:ws}
\end{figure*}

{\small
\bibliographystyle{ieee_fullname}
\bibliography{references}

\begin{thebibliography}{10}\itemsep=-1pt

\bibitem{Akan2021SLAMPSL}
Adil~Kaan Akan, Erkut Erdem, Aykut Erdem, and Fatma Guney.
\newblock Slamp: Stochastic latent appearance and motion prediction.
\newblock {\em 2021 IEEE/CVF International Conference on Computer Vision
  (ICCV)}, pages 14708--14717, 2021.

\bibitem{albergo2022building}
Michael~S Albergo and Eric Vanden-Eijnden.
\newblock Building normalizing flows with stochastic interpolants.
\newblock {\em arXiv preprint arXiv:2209.15571}, 2022.

\bibitem{Aldausari2020VideoGA}
Nuha Aldausari, Arcot Sowmya, Nadine Marcus, and Gelareh Mohammadi.
\newblock Video generative adversarial networks: A review.
\newblock {\em ACM Computing Surveys (CSUR)}, 55:1 -- 25, 2020.

\bibitem{Ba2016LayerN}
Jimmy Ba, Jamie~Ryan Kiros, and Geoffrey~E. Hinton.
\newblock Layer normalization.
\newblock {\em ArXiv}, abs/1607.06450, 2016.

\bibitem{Babaeizadeh2018StochasticVV}
Mohammad Babaeizadeh, Chelsea Finn, D. Erhan, Roy~H. Campbell, and Sergey
  Levine.
\newblock Stochastic variational video prediction.
\newblock {\em ArXiv}, abs/1710.11252, 2018.

\bibitem{babaeizadeh2021fitvid}
Mohammad Babaeizadeh, Mohammad~Taghi Saffar, Suraj Nair, Sergey Levine, Chelsea
  Finn, and Dumitru Erhan.
\newblock Fitvid: Overfitting in pixel-level video prediction.
\newblock {\em arXiv preprint arXiv:2106.13195}, 2021.

\bibitem{bao2022all}
Fan Bao, Chongxuan Li, Yue Cao, and Jun Zhu.
\newblock All are worth words: a vit backbone for score-based diffusion models.
\newblock {\em arXiv preprint arXiv:2209.12152}, 2022.

\bibitem{Brock2019LargeSG}
Andrew Brock, Jeff Donahue, and Karen Simonyan.
\newblock Large scale gan training for high fidelity natural image synthesis.
\newblock {\em ArXiv}, abs/1809.11096, 2019.

\bibitem{Brown2020LanguageMA}
Tom~B. Brown, Benjamin Mann, Nick Ryder, Melanie Subbiah, Jared Kaplan,
  Prafulla Dhariwal, Arvind Neelakantan, Pranav Shyam, Girish Sastry, Amanda
  Askell, Sandhini Agarwal, Ariel Herbert-Voss, Gretchen Krueger, T.~J.
  Henighan, Rewon Child, Aditya Ramesh, Daniel~M. Ziegler, Jeff Wu, Clemens
  Winter, Christopher Hesse, Mark Chen, Eric Sigler, Mateusz Litwin, Scott
  Gray, Benjamin Chess, Jack Clark, Christopher Berner, Sam McCandlish, Alec
  Radford, Ilya Sutskever, and Dario Amodei.
\newblock Language models are few-shot learners.
\newblock {\em ArXiv}, abs/2005.14165, 2020.

\bibitem{Carreira2017QuoVA}
Jo{\~a}o Carreira and Andrew Zisserman.
\newblock Quo vadis, action recognition? a new model and the kinetics dataset.
\newblock {\em 2017 IEEE Conference on Computer Vision and Pattern Recognition
  (CVPR)}, pages 4724--4733, 2017.

\bibitem{castrejon2019improved}
Lluis Castrejon, Nicolas Ballas, and Aaron Courville.
\newblock Improved conditional vrnns for video prediction.
\newblock In {\em Proceedings of the IEEE/CVF International Conference on
  Computer Vision}, pages 7608--7617, 2019.

\bibitem{Chung2022ComeCloserDiffuseFasterAC}
Hyungjin Chung, Byeongsu Sim, and Jong-Chul Ye.
\newblock Come-closer-diffuse-faster: Accelerating conditional diffusion models
  for inverse problems through stochastic contraction.
\newblock {\em 2022 IEEE/CVF Conference on Computer Vision and Pattern
  Recognition (CVPR)}, pages 12403--12412, 2022.

\bibitem{Clark2019AdversarialVG}
Aidan Clark, Jeff Donahue, and Karen Simonyan.
\newblock Adversarial video generation on complex datasets.
\newblock {\em arXiv: Computer Vision and Pattern Recognition}, 2019.

\bibitem{denton2018stochastic}
Emily Denton and Rob Fergus.
\newblock Stochastic video generation with a learned prior.
\newblock In {\em International conference on machine learning}, pages
  1174--1183. PMLR, 2018.

\bibitem{Denton2017UnsupervisedLO}
Emily~L. Denton and Vighnesh Birodkar.
\newblock Unsupervised learning of disentangled representations from video.
\newblock {\em ArXiv}, abs/1705.10915, 2017.

\bibitem{Dhariwal2021DiffusionMB}
Prafulla Dhariwal and Alex Nichol.
\newblock Diffusion models beat gans on image synthesis.
\newblock {\em ArXiv}, abs/2105.05233, 2021.

\bibitem{Dockhorn2022GENIEHD}
Tim Dockhorn, Arash Vahdat, and Karsten Kreis.
\newblock Genie: Higher-order denoising diffusion solvers.
\newblock {\em ArXiv}, abs/2210.05475, 2022.

\bibitem{Dosovitskiy2021AnII}
Alexey Dosovitskiy, Lucas Beyer, Alexander Kolesnikov, Dirk Weissenborn,
  Xiaohua Zhai, Thomas Unterthiner, Mostafa Dehghani, Matthias Minderer, Georg
  Heigold, Sylvain Gelly, Jakob Uszkoreit, and Neil Houlsby.
\newblock An image is worth 16x16 words: Transformers for image recognition at
  scale.
\newblock {\em ArXiv}, abs/2010.11929, 2021.

\bibitem{Ebert2017SelfSupervisedVP}
Frederik Ebert, Chelsea Finn, Alex~X. Lee, and Sergey Levine.
\newblock Self-supervised visual planning with temporal skip connections.
\newblock In {\em CoRL}, 2017.

\bibitem{esser2021taming}
Patrick Esser, Robin Rombach, and Bjorn Ommer.
\newblock Taming transformers for high-resolution image synthesis.
\newblock In {\em Proceedings of the IEEE/CVF conference on computer vision and
  pattern recognition}, pages 12873--12883, 2021.

\bibitem{finn2016unsupervised}
Chelsea Finn, Ian Goodfellow, and Sergey Levine.
\newblock Unsupervised learning for physical interaction through video
  prediction.
\newblock {\em Advances in neural information processing systems}, 29, 2016.

\bibitem{finn2017deep}
Chelsea Finn and Sergey Levine.
\newblock Deep visual foresight for planning robot motion.
\newblock In {\em 2017 IEEE International Conference on Robotics and Automation
  (ICRA)}, pages 2786--2793. IEEE, 2017.

\bibitem{Franceschi2020StochasticLR}
Jean-Yves Franceschi, Edouard Delasalles, Micka{\"e}l Chen, Sylvain Lamprier,
  and Patrick Gallinari.
\newblock Stochastic latent residual video prediction.
\newblock In {\em ICML}, 2020.

\bibitem{Gao2021AccurateGK}
Xiaojie Gao, Yueming Jin, Qi Dou, Chi-Wing Fu, and Pheng-Ann Heng.
\newblock Accurate grid keypoint learning for efficient video prediction.
\newblock {\em 2021 IEEE/RSJ International Conference on Intelligent Robots and
  Systems (IROS)}, pages 5908--5915, 2021.

\bibitem{gao2021accurate}
Xiaojie Gao, Yueming Jin, Qi Dou, Chi-Wing Fu, and Pheng-Ann Heng.
\newblock Accurate grid keypoint learning for efficient video prediction.
\newblock In {\em 2021 IEEE/RSJ International Conference on Intelligent Robots
  and Systems (IROS)}, pages 5908--5915. IEEE, 2021.

\bibitem{Goodfellow2014GenerativeAN}
Ian~J. Goodfellow, Jean Pouget-Abadie, Mehdi Mirza, Bing Xu, David
  Warde-Farley, Sherjil Ozair, Aaron~C. Courville, and Yoshua Bengio.
\newblock Generative adversarial nets.
\newblock In {\em NIPS}, 2014.

\bibitem{gupta2022maskvit}
Agrim Gupta, Stephen Tian, Yunzhi Zhang, Jiajun Wu, Roberto
  Mart{\'\i}n-Mart{\'\i}n, and Li Fei-Fei.
\newblock Maskvit: Masked visual pre-training for video prediction.
\newblock {\em arXiv preprint arXiv:2206.11894}, 2022.

\bibitem{harvey2022flexible}
William Harvey, Saeid Naderiparizi, Vaden Masrani, Christian Weilbach, and
  Frank Wood.
\newblock Flexible diffusion modeling of long videos.
\newblock {\em arXiv preprint arXiv:2205.11495}, 2022.

\bibitem{Hendrycks2016GaussianEL}
Dan Hendrycks and Kevin Gimpel.
\newblock Gaussian error linear units (gelus).
\newblock {\em arXiv: Learning}, 2016.

\bibitem{ho2022imagen}
Jonathan Ho, William Chan, Chitwan Saharia, Jay Whang, Ruiqi Gao, Alexey
  Gritsenko, Diederik~P Kingma, Ben Poole, Mohammad Norouzi, David~J Fleet,
  et~al.
\newblock Imagen video: High definition video generation with diffusion models.
\newblock {\em arXiv preprint arXiv:2210.02303}, 2022.

\bibitem{Ho2020DenoisingDP}
Jonathan Ho, Ajay Jain, and P. Abbeel.
\newblock Denoising diffusion probabilistic models.
\newblock {\em ArXiv}, abs/2006.11239, 2020.

\bibitem{ho2022classifier}
Jonathan Ho and Tim Salimans.
\newblock Classifier-free diffusion guidance.
\newblock {\em arXiv preprint arXiv:2207.12598}, 2022.

\bibitem{ho2022video}
Jonathan Ho, Tim Salimans, Alexey Gritsenko, William Chan, Mohammad Norouzi,
  and David~J Fleet.
\newblock Video diffusion models.
\newblock {\em arXiv preprint arXiv:2204.03458}, 2022.

\bibitem{10.1162/neco.1997.9.8.1735}
Sepp Hochreiter and J\"{u}rgen Schmidhuber.
\newblock Long short-term memory.
\newblock {\em Neural Comput.}, 9(8):1735–1780, nov 1997.

\bibitem{hoppe2022diffusion}
Tobias H{\"o}ppe, Arash Mehrjou, Stefan Bauer, Didrik Nielsen, and Andrea
  Dittadi.
\newblock Diffusion models for video prediction and infilling.
\newblock {\em arXiv preprint arXiv:2206.07696}, 2022.

\bibitem{JolicoeurMartineau2021GottaGF}
Alexia Jolicoeur-Martineau, Ke Li, Remi Piche-Taillefer, Tal Kachman, and
  Ioannis Mitliagkas.
\newblock Gotta go fast when generating data with score-based models.
\newblock {\em ArXiv}, abs/2105.14080, 2021.

\bibitem{Kim2019UnsupervisedKL}
Yunji Kim, Seonghyeon Nam, I. Cho, and Seon~Joo Kim.
\newblock Unsupervised keypoint learning for guiding class-conditional video
  prediction.
\newblock {\em ArXiv}, abs/1910.02027, 2019.

\bibitem{Kingma2014AutoEncodingVB}
Diederik~P. Kingma and Max Welling.
\newblock Auto-encoding variational bayes.
\newblock {\em CoRR}, abs/1312.6114, 2014.

\bibitem{Kong2021OnFS}
Zhifeng Kong and Wei Ping.
\newblock On fast sampling of diffusion probabilistic models.
\newblock {\em ArXiv}, abs/2106.00132, 2021.

\bibitem{Kumar2020VideoFlowAC}
Manoj Kumar, Mohammad Babaeizadeh, D. Erhan, Chelsea Finn, Sergey Levine,
  Laurent Dinh, and Durk Kingma.
\newblock Videoflow: A conditional flow-based model for stochastic video
  generation.
\newblock {\em arXiv: Computer Vision and Pattern Recognition}, 2020.

\bibitem{le2021ccvs}
Guillaume Le~Moing, Jean Ponce, and Cordelia Schmid.
\newblock Ccvs: Context-aware controllable video synthesis.
\newblock {\em Advances in Neural Information Processing Systems},
  34:14042--14055, 2021.

\bibitem{lee2018stochastic}
Alex~X Lee, Richard Zhang, Frederik Ebert, Pieter Abbeel, Chelsea Finn, and
  Sergey Levine.
\newblock Stochastic adversarial video prediction.
\newblock {\em arXiv preprint arXiv:1804.01523}, 2018.

\bibitem{Lee2021RevisitingHA}
Wonkwang Lee, Whie Jung, Han Zhang, Ting Chen, Jing~Yu Koh, Thomas~E. Huang,
  Hyungsuk Yoon, Honglak Lee, and Seunghoon Hong.
\newblock Revisiting hierarchical approach for persistent long-term video
  prediction.
\newblock {\em ArXiv}, abs/2104.06697, 2021.

\bibitem{Liang2017DualMG}
Xiaodan Liang, Lisa Lee, Wei Dai, and Eric~P. Xing.
\newblock Dual motion gan for future-flow embedded video prediction.
\newblock {\em 2017 IEEE International Conference on Computer Vision (ICCV)},
  pages 1762--1770, 2017.

\bibitem{lipman2022flow}
Yaron Lipman, Ricky~TQ Chen, Heli Ben-Hamu, Maximilian Nickel, and Matt Le.
\newblock Flow matching for generative modeling.
\newblock {\em arXiv preprint arXiv:2210.02747}, 2022.

\bibitem{liu2022flow}
Xingchao Liu, Chengyue Gong, and Qiang Liu.
\newblock Flow straight and fast: Learning to generate and transfer data with
  rectified flow.
\newblock {\em arXiv preprint arXiv:2209.03003}, 2022.

\bibitem{Loshchilov2019DecoupledWD}
Ilya Loshchilov and Frank Hutter.
\newblock Decoupled weight decay regularization.
\newblock In {\em ICLR}, 2019.

\bibitem{luc2020transformation}
Pauline Luc, Aidan Clark, Sander Dieleman, Diego de~Las Casas, Yotam Doron,
  Albin Cassirer, and Karen Simonyan.
\newblock Transformation-based adversarial video prediction on large-scale
  data.
\newblock {\em arXiv preprint arXiv:2003.04035}, 2020.

\bibitem{Luc2020TransformationbasedAV}
Pauline Luc, Aidan Clark, Sander Dieleman, Diego de Las~Casas, Yotam Doron,
  Albin Cassirer, and Karen Simonyan.
\newblock Transformation-based adversarial video prediction on large-scale
  data.
\newblock {\em ArXiv}, abs/2003.04035, 2020.

\bibitem{Marchesi2017MegapixelSI}
Marco Marchesi.
\newblock Megapixel size image creation using generative adversarial networks.
\newblock {\em ArXiv}, abs/1706.00082, 2017.

\bibitem{Mathieu2016DeepMV}
Micha{\"e}l Mathieu, Camille Couprie, and Yann LeCun.
\newblock Deep multi-scale video prediction beyond mean square error.
\newblock {\em CoRR}, abs/1511.05440, 2016.

\bibitem{Mei2022VIDMVI}
Kangfu Mei and Vishal~M. Patel.
\newblock Vidm: Video implicit diffusion models.
\newblock {\em ArXiv}, abs/2212.00235, 2022.

\bibitem{Minderer2019UnsupervisedLO}
Matthias Minderer, Chen Sun, Ruben Villegas, Forrester Cole, Kevin~P. Murphy,
  and Honglak Lee.
\newblock Unsupervised learning of object structure and dynamics from videos.
\newblock {\em ArXiv}, abs/1906.07889, 2019.

\bibitem{Oliu2018FoldedRN}
Marc Oliu, Javier Selva, and Sergio Escalera.
\newblock Folded recurrent neural networks for future video prediction.
\newblock In {\em ECCV}, 2018.

\bibitem{Parmar2022OnAR}
Gaurav Parmar, Richard Zhang, and Junyan Zhu.
\newblock On aliased resizing and surprising subtleties in gan evaluation.
\newblock {\em 2022 IEEE/CVF Conference on Computer Vision and Pattern
  Recognition (CVPR)}, pages 11400--11410, 2022.

\bibitem{rakhimov2020latent}
Ruslan Rakhimov, Denis Volkhonskiy, Alexey Artemov, Denis Zorin, and Evgeny
  Burnaev.
\newblock Latent video transformer.
\newblock {\em arXiv preprint arXiv:2006.10704}, 2020.

\bibitem{Ramesh2022HierarchicalTI}
Aditya Ramesh, Prafulla Dhariwal, Alex Nichol, Casey Chu, and Mark Chen.
\newblock Hierarchical text-conditional image generation with clip latents.
\newblock {\em ArXiv}, abs/2204.06125, 2022.

\bibitem{rombach2022high}
Robin Rombach, Andreas Blattmann, Dominik Lorenz, Patrick Esser, and Bj{\"o}rn
  Ommer.
\newblock High-resolution image synthesis with latent diffusion models.
\newblock In {\em Proceedings of the IEEE/CVF Conference on Computer Vision and
  Pattern Recognition}, pages 10684--10695, 2022.

\bibitem{Ronneberger2015UNetCN}
Olaf Ronneberger, Philipp Fischer, and Thomas Brox.
\newblock U-net: Convolutional networks for biomedical image segmentation.
\newblock {\em ArXiv}, abs/1505.04597, 2015.

\bibitem{schuldt2004recognizing}
Christian Schuldt, Ivan Laptev, and Barbara Caputo.
\newblock Recognizing human actions: a local svm approach.
\newblock In {\em Proceedings of the 17th International Conference on Pattern
  Recognition, 2004. ICPR 2004.}, volume~3, pages 32--36. IEEE, 2004.

\bibitem{seo2021autoregressive}
Younggyo Seo, Kimin Lee, Fangchen Liu, Stephen James, and P. Abbeel.
\newblock Harp: Autoregressive latent video prediction with high-fidelity image
  generator.
\newblock {\em ArXiv}, abs/2209.07143, 2022.

\bibitem{Skorokhodov2021StyleGANVAC}
Ivan Skorokhodov, S. Tulyakov, and Mohamed Elhoseiny.
\newblock Stylegan-v: A continuous video generator with the price, image
  quality and perks of stylegan2.
\newblock {\em 2022 IEEE/CVF Conference on Computer Vision and Pattern
  Recognition (CVPR)}, pages 3616--3626, 2021.

\bibitem{Song2021ScoreBasedGM}
Yang Song, Jascha~Narain Sohl-Dickstein, Diederik~P. Kingma, Abhishek Kumar,
  Stefano Ermon, and Ben Poole.
\newblock Score-based generative modeling through stochastic differential
  equations.
\newblock {\em ArXiv}, abs/2011.13456, 2021.

\bibitem{Sun2018ATV}
Ximeng Sun, Huijuan Xu, and Kate Saenko.
\newblock A two-stream variational adversarial network for video generation.
\newblock {\em ArXiv}, abs/1812.01037, 2018.

\bibitem{Tashiro2021CSDICS}
Yusuke Tashiro, Jiaming Song, Yang Song, and Stefano Ermon.
\newblock Csdi: Conditional score-based diffusion models for probabilistic time
  series imputation.
\newblock In {\em NeurIPS}, 2021.

\bibitem{Tulyakov2018MoCoGANDM}
S. Tulyakov, Ming-Yu Liu, Xiaodong Yang, and Jan Kautz.
\newblock Mocogan: Decomposing motion and content for video generation.
\newblock {\em 2018 IEEE/CVF Conference on Computer Vision and Pattern
  Recognition}, pages 1526--1535, 2018.

\bibitem{Unterthiner2018TowardsAG}
Thomas Unterthiner, Sjoerd van Steenkiste, Karol Kurach, Rapha{\"e}l Marinier,
  Marcin Michalski, and Sylvain Gelly.
\newblock Towards accurate generative models of video: A new metric \&
  challenges.
\newblock {\em ArXiv}, abs/1812.01717, 2018.

\bibitem{Oord2017NeuralDR}
A{\"a}ron van~den Oord, Oriol Vinyals, and Koray Kavukcuoglu.
\newblock Neural discrete representation learning.
\newblock In {\em NIPS}, 2017.

\bibitem{NIPS2017_3f5ee243}
Ashish Vaswani, Noam Shazeer, Niki Parmar, Jakob Uszkoreit, Llion Jones,
  Aidan~N Gomez, \L~ukasz Kaiser, and Illia Polosukhin.
\newblock Attention is all you need.
\newblock In I. Guyon, U.~Von Luxburg, S. Bengio, H. Wallach, R. Fergus, S.
  Vishwanathan, and R. Garnett, editors, {\em Advances in Neural Information
  Processing Systems}, volume~30. Curran Associates, Inc., 2017.

\bibitem{Villegas2019HighFV}
Ruben Villegas, Arkanath Pathak, Harini Kannan, D. Erhan, Quoc~V. Le, and
  Honglak Lee.
\newblock High fidelity video prediction with large stochastic recurrent neural
  networks.
\newblock {\em ArXiv}, abs/1911.01655, 2019.

\bibitem{Villegas2017DecomposingMA}
Ruben Villegas, Jimei Yang, Seunghoon Hong, Xunyu Lin, and Honglak Lee.
\newblock Decomposing motion and content for natural video sequence prediction.
\newblock {\em ArXiv}, abs/1706.08033, 2017.

\bibitem{voleti2022mcvd}
Vikram Voleti, Alexia Jolicoeur-Martineau, and Christopher Pal.
\newblock Mcvd: Masked conditional video diffusion for prediction, generation,
  and interpolation.
\newblock {\em arXiv preprint arXiv:2205.09853}, 2022.

\bibitem{wang2018predrnn++}
Yunbo Wang, Zhifeng Gao, Mingsheng Long, Jianmin Wang, and S~Yu Philip.
\newblock Predrnn++: Towards a resolution of the deep-in-time dilemma in
  spatiotemporal predictive learning.
\newblock In {\em International Conference on Machine Learning}, pages
  5123--5132. PMLR, 2018.

\bibitem{Wang2004ImageQA}
Zhou Wang, Alan~Conrad Bovik, Hamid~R. Sheikh, and Eero~P. Simoncelli.
\newblock Image quality assessment: from error visibility to structural
  similarity.
\newblock {\em IEEE Transactions on Image Processing}, 13:600--612, 2004.

\bibitem{watson2022novel}
Daniel Watson, William Chan, Ricardo Martin-Brualla, Jonathan Ho, Andrea
  Tagliasacchi, and Mohammad Norouzi.
\newblock Novel view synthesis with diffusion models.
\newblock {\em arXiv preprint arXiv:2210.04628}, 2022.

\bibitem{weissenborn2019scaling}
Dirk Weissenborn, Oscar T{\"a}ckstr{\"o}m, and Jakob Uszkoreit.
\newblock Scaling autoregressive video models.
\newblock {\em arXiv preprint arXiv:1906.02634}, 2019.

\bibitem{Wichers2018HierarchicalLV}
Nevan Wichers, Ruben Villegas, D. Erhan, and Honglak Lee.
\newblock Hierarchical long-term video prediction without supervision.
\newblock {\em ArXiv}, abs/1806.04768, 2018.

\bibitem{Wu2022NWAVS}
Chenfei Wu, Jian Liang, Lei Ji, F. Yang, Yuejian Fang, Daxin Jiang, and Nan
  Duan.
\newblock N{\"u}wa: Visual synthesis pre-training for neural visual world
  creation.
\newblock In {\em ECCV}, 2022.

\bibitem{xie2021learning}
Annie Xie, Dylan Losey, Ryan Tolsma, Chelsea Finn, and Dorsa Sadigh.
\newblock Learning latent representations to influence multi-agent interaction.
\newblock In {\em Conference on robot learning}, pages 575--588. PMLR, 2021.

\bibitem{yan2021videogpt}
Wilson Yan, Yunzhi Zhang, Pieter Abbeel, and Aravind Srinivas.
\newblock Videogpt: Video generation using vq-vae and transformers.
\newblock {\em arXiv preprint arXiv:2104.10157}, 2021.

\bibitem{yang2022diffusion}
Ruihan Yang, Prakhar Srivastava, and Stephan Mandt.
\newblock Diffusion probabilistic modeling for video generation.
\newblock {\em arXiv preprint arXiv:2203.09481}, 2022.

\bibitem{Yi2020CLEVRERCE}
Kexin Yi, Chuang Gan, Yunzhu Li, Pushmeet Kohli, Jiajun Wu, Antonio Torralba,
  and Joshua~B. Tenenbaum.
\newblock Clevrer: Collision events for video representation and reasoning.
\newblock {\em ArXiv}, abs/1910.01442, 2020.

\bibitem{Yu2022GeneratingVW}
Sihyun Yu, Jihoon Tack, Sangwoo Mo, Hyunsu Kim, Junho Kim, Jung-Woo Ha, and
  Jinwoo Shin.
\newblock Generating videos with dynamics-aware implicit generative adversarial
  networks.
\newblock {\em ArXiv}, abs/2202.10571, 2022.

\bibitem{Zhang2018ImageSU}
Yulun Zhang, Kunpeng Li, Kai Li, Lichen Wang, Bineng Zhong, and Yun~Raymond Fu.
\newblock Image super-resolution using very deep residual channel attention
  networks.
\newblock In {\em European Conference on Computer Vision}, 2018.

\end{thebibliography}
}

\end{document}